\PassOptionsToPackage{table}{xcolor}
\documentclass{jingdong}

\usepackage[utf8]{inputenc}
\usepackage[T1]{fontenc}
\usepackage{float}
\usepackage{hyperref}
\usepackage{url}
\usepackage{booktabs}
\usepackage{amsmath}
\usepackage{amssymb}
\usepackage{amsfonts}
\usepackage{nicefrac}
\usepackage{microtype}
\usepackage[table]{xcolor}
\usepackage{graphicx}
\usepackage{multirow}
\usepackage{makecell}
\usepackage{placeins}
\usepackage{longtable}
\usepackage{etoc}

\let\cite\citep

\newcommand{\ours}{OneStreamer}
\title{\texorpdfstring{\mbox{\raisebox{-0.2em}[0pt][0pt]{\includegraphics[height=1.2em]{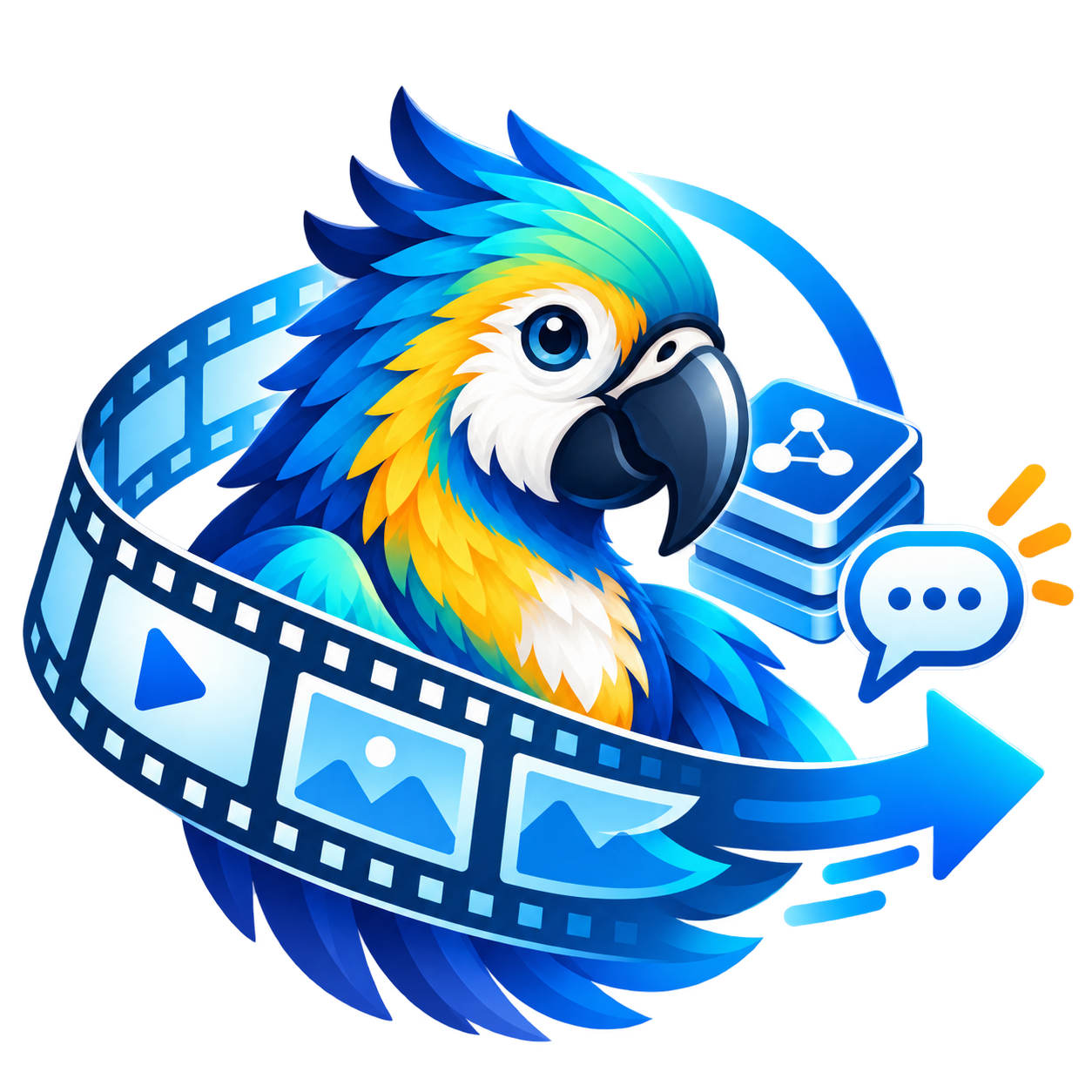}}\hspace{0.25em}\onestreamerwordmark}}{OneStreamer}: Unifying Perception, Memory, and Proactive Response in Streaming Video Interaction}

\renewcommand{\authorfont}{\fontsize{9.5}{11.5}\selectfont}
\renewcommand{\affiliationfont}{\fontsize{9}{11}\selectfont}
\renewcommand{\contributionfont}{\fontsize{9}{11}\selectfont}

\author[*\,1]{Xiangyu Zeng}
\author[*\,1]{Yuandong Yang}
\author[*\,2,4]{Zhiqiu Zhang}
\author[*\,1]{Yuhan Zhu}
\author[*\,1]{Xinhao Li}
\author[*\,3]{Qingyi Si}
\author[6]{Dingyu Yao}
\author[1,2]{Changlian Ma}
\author[1,2]{Haoran Chen}
\author[1,2]{Xinyu Chen}
\author[2,5]{Yansong Shi}
\author[6]{Junhao Zhou}
\author[9]{Yifei Li}
\author[1,3]{Jun Zhang}
\author[6]{Chuanyu Qin}
\author[6]{Chenxu Yang}
\author[7]{Xinlei Yu}
\author[8]{Kun Ouyang}
\author[6]{Yuchen Shao}
\author[6]{Qianshan Wei}
\author[10]{Changhai Zhou}
\author[11]{Jun Gao}
\author[3]{Jiaqi Wang}
\author[\dagger\,1]{Limin Wang}

\affiliation[1]{NJU}
\affiliation[2]{PJLAB}
\affiliation[3]{JD}
\affiliation[4]{SJTU}
\affiliation[5]{USTC}
\affiliation[6]{CAS}
\affiliation[7]{CUHK}
\affiliation[8]{PKU}
\affiliation[9]{THU}
\affiliation[10]{FDU}
\affiliation[11]{ZJU}

\contribution[*]{Equal contribution}
\contribution[\dagger]{Corresponding author}
\checkdata[Project homepage]{\href{https://mcg-nju.github.io/OneStreamer}{\textcolor{onestreamerblue}{\nolinkurl{https://mcg-nju.github.io/OneStreamer}}}}

\abstract{
Streaming video LLMs must retain evidence before its relevance to future tasks is known and respond when sufficient evidence becomes available.
The challenge is to form reusable factual memory without compromising real-time perception.
We introduce \textbf{{\ours}}, which jointly learns query-independent evidence recording and task response through a shared proactive generation process.
Its \textbf{Proactive Hierarchical Caption Memory (PHCM)} produces time-grounded local-detail captions and summaries of completed events.
Streaming caption targets supervise the interpretation of observed video prefixes during training.
At inference, model-generated records complement a recent visual window, providing reusable factual context without revisiting historical visual features.
\textbf{Proactive State Transition Learning (PSTL)} reduces the dominance of repeated waiting states by preserving supervision at all output anchors and selecting representative state-change and state-persistence tokens.
We further develop a streaming data synthesis pipeline that aligns output content and timing with available evidence.
Combining the resulting streaming captions and QA with cleaned open-source data yields \textbf{OneStreamer-1M}, a broad-coverage streaming video interaction dataset with over one million records spanning diverse tasks.
Our 4B model achieves the best results among the compared methods across all eight evaluated streaming video understanding benchmarks.
Ablations show that retaining generated captions improves historical QA without degrading real-time perception.
PSTL also outperforms dense state supervision while supervising only 27.5\% of annotated state tokens.
Together, these results support proactive generation as a shared learning interface connecting perception, memory formation, and timely response in streaming video interaction.
}

\begin{document}
\etocdepthtag.toc{main}

\maketitle

\section{Introduction}

Real-world video arrives as an open-ended stream of observations. In live-stream assistants, wearable agents, and security monitoring systems, a model must interpret incoming visual evidence without knowing which observations will matter to future questions or tasks. An event that initially appears unimportant may become relevant to a later task after its source frames have left the model's limited visual context. At the same time, a user-visible response must be grounded in the evidence available so far and delivered before the relevant moment passes~\citep{chen2024videollmonline,huang2025videochatonline,qian2025dispider}. We therefore ask: \emph{How can a streaming video LLM turn incoming observations into reusable factual memory for timely responses that draw on both past and current evidence?}

Despite recent advances in continual perception, long-term memory, and response control~\citep{xia2026streamo,xie2026streamov}, streaming video LLMs still face limitations in how they retain evidence and learn to act on it.
Memory mechanisms that compress or retrieve distant visual evidence extend the accessible history~\citep{zhang2024flashvstream,wang2026curvestream}, but retained historical visual tokens can compete with recent observations for limited context capacity, potentially weakening real-time perception~\citep{shen2026simplestream}.
Streaming thinking methods maintain evolving reasoning~\citep{zhang2026thinkasyousee,guan2026videostreamingthinking,liu2026thinkingstreamingvideo}, yet their reasoning traces do not necessarily serve as reusable time-grounded factual records for future tasks.
State-token-based response control enables proactive outputs~\citep{chen2024videollmonline,xia2026streamo}, but dense supervision of these tokens can overemphasize repeated silence labels relative to sparse response decisions.
Together, these limitations highlight a central challenge: forming reusable factual memory without compromising real-time perception, while learning when the available evidence warrants a response.

We address this challenge by rethinking the role of proactive generation: beyond producing user-visible responses, a streaming model can learn to turn observed evidence into reusable factual records, connecting perception learning with query-independent memory formation.
From this perspective, we introduce \textbf{{\ours}}, a streaming video LLM that unifies evidence recording and task response through a shared generation process.
Building on generative response control~\citep{xia2026streamo}, it jointly learns when to record or respond and what to generate by predicting task-specific control tokens and associated text outputs within a single causal visual--language sequence.

\begin{figure*}[t]
    \centering
    \includegraphics[width=\textwidth]{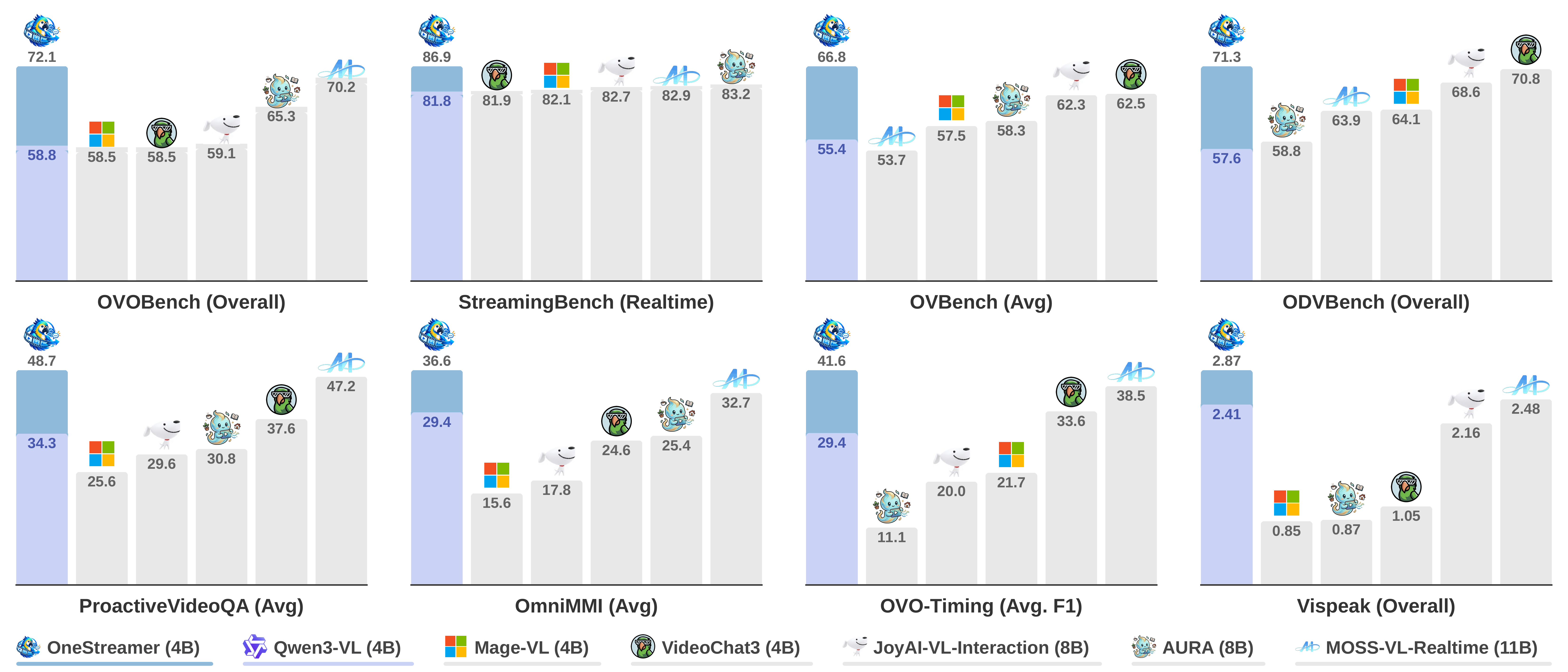}
    \caption{
        \textbf{Evaluation results of OneStreamer.}
        Blue bars highlight OneStreamer, with light purple portions indicating the Qwen3-VL baseline.
        OneStreamer achieves state-of-the-art results across all eight benchmarks, with an average relative improvement of 25.0\% over the Qwen3-VL baseline and 6.1\% over the strongest competing method on each benchmark.
    }
    \label{fig:benchmark_overview}
\end{figure*}

Within this formulation, \textbf{Proactive Hierarchical Caption Memory} organizes observed evidence into time-aligned textual records at two complementary scales.
It generates dense local-detail captions from already observed visual content and sparser semantic summaries of completed events.
During training, causal streaming caption targets supervise the interpretation of observed video prefixes.
At inference, accumulated model-generated records complement a recent visual window, providing distant factual context alongside current visual detail without retrieving or revisiting historical visual features.
To learn when to record or respond, \textbf{Proactive State Transition Learning} addresses the imbalance in state supervision between repeated waiting states and sparse output decisions.
It preserves supervision at all output anchors and selects representative state-change and state-persistence positions.
Unselected state positions are masked only from the state loss, leaving the full sequence and supervision of all task text unchanged.
Together, these designs connect perception supervision, memory reuse, and response timing within the same causal generation process.

Learning this shared process requires more than causal inputs: each target must be supported by evidence available at its assigned time.
Thus, we develop a reusable streaming data synthesis pipeline that transforms offline video resources into streaming training sequences.
The caption branch converts verified multi-granularity annotations into streaming caption targets with evidence-aligned release times.
The QA branch produces evidence-grounded streaming QA sequences with response times calibrated to when sufficient evidence becomes available.
Combining these examples with cleaned open-source data in a common streaming format yields \textbf{OneStreamer-1M}, a broad-coverage training corpus for streaming video interaction.

Our 4B {\ours} model achieves the best results across all eight evaluated online video understanding and proactive-response benchmarks.
Controlled ablations provide three complementary findings.
First, retaining proactively generated captions improves historical QA after their source frames leave the recent visual window.
Second, combining recent visual context with distant caption memory provides a better perception--memory balance than retaining the full visual history or only recent frames.
Third, PSTL outperforms dense state-token supervision and a supervision-matched random baseline on proactive-response benchmarks while supervising only 27.5\% of annotated state tokens.
Together, these results support proactive generation as a shared mechanism connecting visual perception, reusable factual memory, and timely task response within a single streaming model.

\section{Related Work}

\begin{figure}[t]
    \centering
    \includegraphics[width=0.9\textwidth]{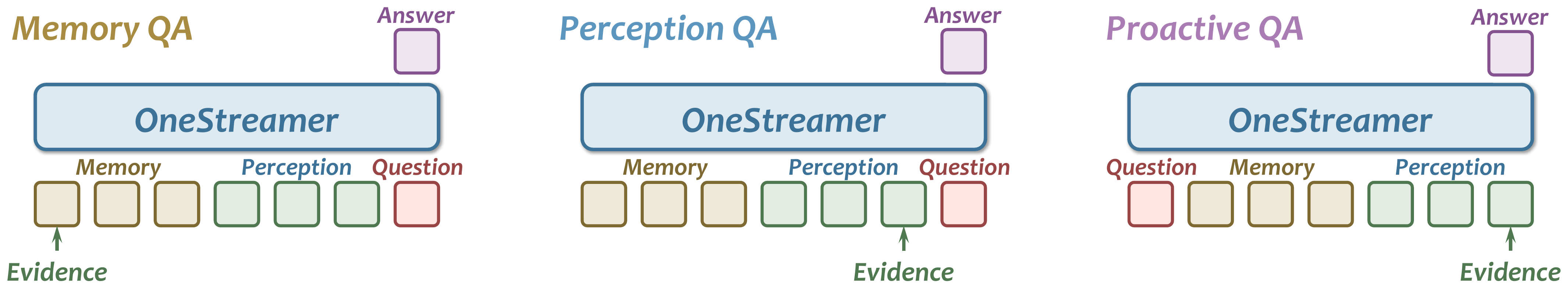}
    \caption{\textbf{Three QA settings in streaming video understanding.} Memory QA answers using past evidence retained in memory; Perception QA answers from current visual evidence; Proactive QA receives the question first and responds once sufficient evidence arrives.}
    \label{fig:three_task_qa}
\end{figure}

\subsection{Perception and Memory in Video Understanding}

Offline video LLMs handle long recordings through memory compression~\citep{song2024moviechat,li2026videochatflash,zeng2025timesuite,wang2025internvideo25}, context extension~\citep{zhang2024longva}, or adaptive token selection~\citep{shen2024longvu}.
Online assistants must instead retain evidence from observed prefixes even when its relevance to future queries is unknown.
Existing systems use structured or hierarchical memory~\citep{zhang2024flashvstream,huang2025videochatonline,zeng2026streamforest,xie2026fluxmem}, budgeted retrieval~\citep{xie2026streamov}, recent windows~\citep{xu2026streamingvlm}, or informative-frame selection~\citep{yao2025timechatonline,wang2026curvestream}.
Under limited context capacity, retaining history can compete with current visual detail, while strict recent windows discard distant events~\citep{shen2026simplestream}.
Our approach complements recent visual inputs with proactively generated, time-grounded factual records, separating query-independent memory writing from later task-conditioned use.

\subsection{Streaming Description and Proactive Response}

Streaming captioning and online video-language modeling support causal frame--text generation~\citep{zhou2024streamingdvc,chen2024videollmonline}, while continuous-commentary methods maintain narration as observations arrive~\citep{chen2025livecc,yang2026livestar,yan2026proactvl}.
Streaming thinking supports online understanding through evolving reasoning or reasoning-oriented memory~\citep{zhang2026thinkasyousee,guan2026videostreamingthinking,liu2026thinkingstreamingvideo,wang2026thinkwhilewatching}.
Proactive methods model response timing through generative state tokens~\citep{xia2026streamo,lu2026aura}, auxiliary information, relevance, or readiness signals~\citep{qian2025dispider,wang2026streambridge,azad2026streamready,xie2026streamov,li2025lionfs,ding2025streammind}, or policies learned through anticipatory planning and reinforcement learning~\citep{wang2026mmduet2,yang2025streamagent}.
Reasoning traces do not necessarily constitute reusable factual records, and relevant evidence does not necessarily warrant a user-visible response.
{\ours} jointly models factual recording and task response, using selective state supervision that preserves all output anchors and representative state-change and state-persistence positions.

\section{Method}

In this section, we detail how {\ours} unifies query-independent evidence recording and task response through a shared causal generation process that jointly models when to record or respond and what to generate.

\subsection{Overall Architecture}

\begin{figure*}[t]
    \centering
    \includegraphics[width=\textwidth]{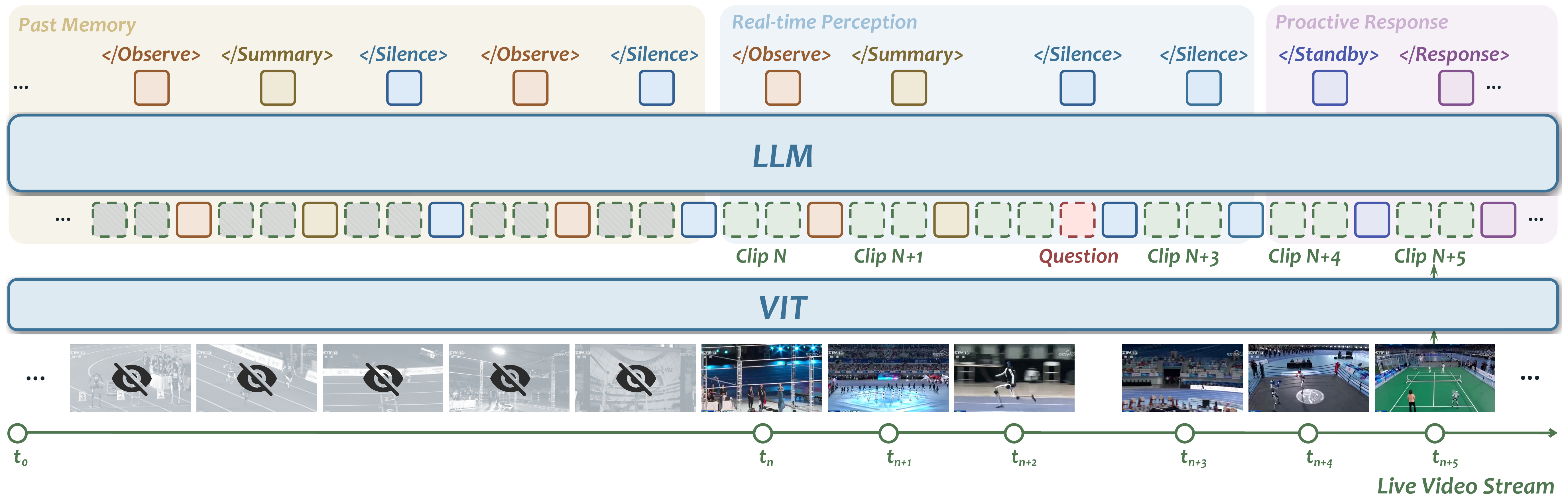} 
    \caption{\textbf{Architecture of {\ours}.} Recent visual tokens are interleaved with local-detail captions \texttt{</Observe>} and semantic summaries \texttt{</Summary>} retained as caption memory. For proactive QA, \texttt{</Silence>}, \texttt{</Standby>}, and \texttt{</Response>} denote continued observation, emerging but insufficient evidence, and response generation, respectively.}
    \label{fig:architecture}
\end{figure*}

As shown in Figure~\ref{fig:architecture}, {\ours} processes a live video stream incrementally as temporally ordered clips. A vision encoder extracts visual tokens from each incoming clip and a projector maps them into the LLM's embedding space. A Recent-$N$ FIFO sliding window retains only the visual tokens corresponding to the latest $N$ observed frames. These tokens are interleaved with a text history that includes accumulated caption records and dialogue. The resulting causal visual-language sequence contains only information available at the current time. The LLM predicts task-specific control tokens from this sequence and generates the associated text when the predicted state initiates an output.

The model uses task-specific control tokens to indicate whether to record evidence, continue observing, or produce a user-visible response. For memory formation, \texttt{</Observe>} introduces a local-detail caption of observed objects, actions, scenes, and state changes. \texttt{</Summary>} introduces a semantic summary of a completed event or segment. Both caption types are aligned with their source intervals and appended to the text history as factual context for subsequent predictions.  Proactive QA uses \texttt{</Standby>} to indicate that relevant evidence is emerging but the model is not yet ready to answer. The model generates user-visible answers or other task outputs only after \texttt{</Response>}. The shared \texttt{</Silence>} token indicates continued observation without generating caption or response.

\subsection{Proactive Hierarchical Caption Memory}

Under limited context capacity, retaining historical visual tokens can compete with the fine-grained visual evidence needed for current perception, whereas a Recent-$N$ window alone discards distant visual history. Therefore, we introduce Proactive Hierarchical Caption Memory (PHCM), which complements recent visual tokens with time-aligned textual captions. As the stream unfolds, {\ours} proactively generates these records from already observed content and retains them after their source frames leave the visual window. This heterogeneous representation extends access to past events through compact text while keeping the Recent-$N$ visual window unchanged.

PHCM organizes observed evidence into two types of time-aligned records, each introduced by a dedicated control token. \texttt{</Observe>} introduces dense local-detail captions describing directly observed objects, actions, scenes, and state changes over short intervals. \texttt{</Summary>} introduces sparser semantic summaries of completed events or segments at a coarser temporal granularity. Each record is associated with its source interval, while the two granularities form a hierarchy of local details and event-level summaries. Both record types are supervised to describe only content supported by the video observed so far, providing reusable factual context for subsequent predictions.

At inference time, {\ours} incrementally builds PHCM using only the currently available visual-language context. Outside user-visible answer generation, a predicted \texttt{</Observe>} or \texttt{</Summary>} initiates the corresponding record, whereas \texttt{</Silence>} continues observation without writing one. Each generated record is appended to the text history in generation order and becomes available to subsequent predictions. The resulting text history, including all accumulated records and dialogue, is interleaved with the visual tokens retained in the Recent-$N$ window, providing long-range context without retrieving or revisiting historical visual features.

This design connects perception learning with query-independent memory formation. During training, causal streaming caption targets supervise the interpretation of observed video prefixes. At inference, the model generates and retains time-grounded records as explicit context for subsequent predictions, turning streaming description into reusable factual memory.

\subsection{Proactive State Transition Learning}

Proactive streaming interaction requires the model to decide when the available evidence warrants a memory record or a task response. Dense state-token supervision can overemphasize waiting when repeated silence tokens outnumber output-initiating tokens, yet supervising only state changes omits direct supervision of when the current state should persist. To address this, we introduce \textbf{Proactive State Transition Learning (PSTL)} to preserve supervision at all output anchors and select representative state tokens associated with state changes or persistence. Unselected state tokens remain in the causal sequence but are excluded from the state loss.

We implement PSTL through selective state-token supervision guided by task-specific output anchors. These anchors are control states that initiate textual outputs. Within each training sequence, we group state tokens by the transition from the preceding control state to the current one. The size of the largest group targeting an output anchor defines a supervision quota shared across all groups in that sequence. We retain full supervision for groups within the quota and uniformly subsample larger groups without replacement to match the quota. This preserves supervision for all output-initiating tokens while selecting examples of both state changes and state persistence. Unselected state tokens remain in the causal sequence but are excluded from the state loss.

\begin{figure}[t]
    \centering
    \includegraphics[width=\textwidth]{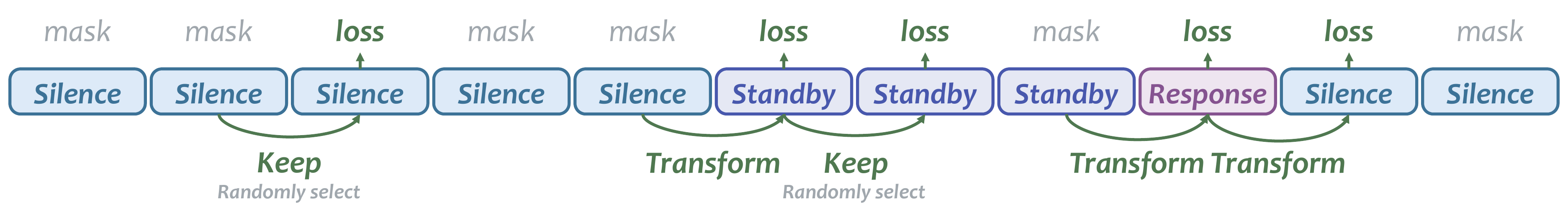}
    \caption{\textbf{Illustration of PSTL state-token selection.} PSTL retains selected tokens for state changes and state persistence, while unselected state tokens are masked only from the state loss. All control tokens remain in the causal sequence.}
    \label{fig:state_transition}
\end{figure}

Formally, let $\mathcal{A}_{\tau}$ denote the output-anchor set for task $\tau$. Let $n_{X \rightarrow Y}$ be the number of occurrences of transition $X \rightarrow Y$ between consecutive annotated control states in a training sequence. Here $X$ and $Y$ range over the task's control states. We define a supervision quota shared across all transition groups in this sequence:
\begin{equation}
    q = \max_X \max_{a \in \mathcal{A}_{\tau}} n_{X \rightarrow a}.
    \label{eq:transition_quota}
\end{equation}
From each transition group $X \rightarrow Y$, we sample state-token indices uniformly without replacement to form $\mathcal{I}_{X \rightarrow Y}$ of size
\begin{equation}
    |\mathcal{I}_{X \rightarrow Y}| = \min(n_{X \rightarrow Y}, q).
    \label{eq:transition_sampling}
\end{equation}
Only state tokens indexed by $\mathcal{I}=\bigcup_{X,Y}\mathcal{I}_{X \rightarrow Y}$ contribute to the state loss. Appendix~\ref{sec:app:pstl} specifies the task-specific anchor sets and the loss function.

By preserving all output anchors and capping frequent transition groups, PSTL limits the influence of repeated silence tokens on the state objective. Masking affects only the state loss. All control tokens remain in the causal training sequence, while captions and task outputs retain full supervision. At inference, the model generates control states and text directly from its learned conditional distributions without applying PSTL or an external transition policy.

\FloatBarrier

\section{Dataset Construction}

\begin{figure}[t]
    \centering
    \includegraphics[width=\textwidth]{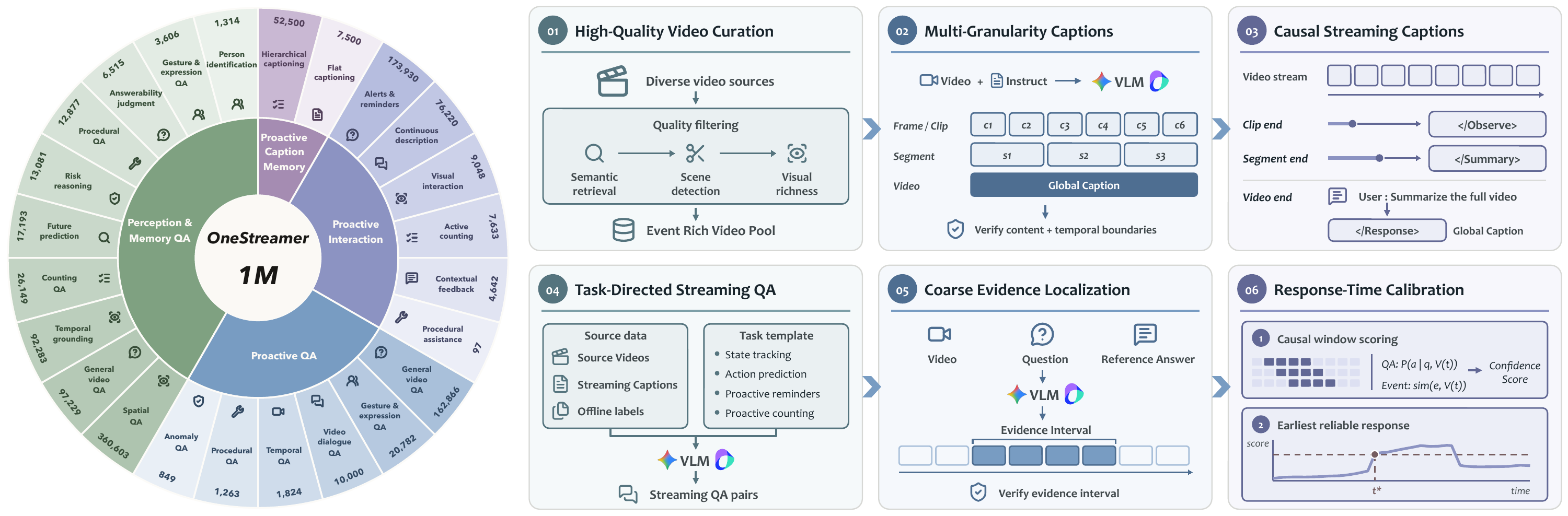}
    \caption{\textbf{Streaming data synthesis pipeline and OneStreamer-1M task composition.} Left: task families and per-task record counts of OneStreamer-1M. Right: stages 01-03 produce verified streaming captions with evidence-aligned release times, while stages 04-06 produce evidence-grounded streaming QA sequences with calibrated response times.}
    \label{fig:data_pipeline}
\end{figure}

High-quality open-source training data that jointly align visual evidence, output content, and response timing remain scarce.
We therefore develop a reusable streaming data synthesis pipeline with two complementary branches: streaming caption synthesis and streaming QA synthesis.
Combining the synthesized examples with cleaned open-source data yields \textbf{OneStreamer-1M}, a broad-coverage training dataset for streaming video interaction.
Figure~\ref{fig:data_pipeline} summarizes the pipeline and the dataset's task composition.
Appendix~\ref{sec:app:dataset} provides further technical details on streaming caption and QA construction, together with data sources, statistics, and representative caption construction examples.

\subsection{Streaming Caption Synthesis}

Offline video captions require adaptation for streaming training, where each output must be grounded in the observed video prefix.
The caption branch aligns both caption content and release time with the available evidence.
Local descriptions are released only after their supporting visual intervals have been observed, whereas segment-level summaries are released only after the corresponding events or segments are complete.
This branch consists of the following three stages.

\textbf{\textit{High-quality video curation.}}
We first construct a diverse candidate pool using semantic retrieval, scene detection, and visual-richness assessment, prioritizing videos with clear visual changes, coherent temporal structure, and rich event content. We filter out videos with low visual quality, prolonged static periods, black frames, excessive shot fragmentation, or decoding failures.
\textbf{\textit{Multi-granularity caption annotation.}}
For each selected video, we use Gemini and Seed to generate timestamp-grounded captions at frame/clip, segment, and video levels. Frame/clip captions capture local actions and events, segment captions summarize coherent events across clips, and video captions capture cross-segment relations and global event structure. We verify all levels against the source video for factual and temporal consistency, discarding unsupported or temporally misaligned captions.
\textbf{\textit{Causal streaming sequence construction.}}
We convert the verified annotations into streaming visual-language sequences with evidence-aligned output timing. Frame/clip captions become \texttt{</Observe>} targets at the end of their supporting intervals, while segment captions become \texttt{</Summary>} targets at segment boundaries. After the stream ends, we append a full-video summary instruction and use the video-level caption as the answer target following \texttt{</Response>}. Each sequence combines dense local records, sparse segment summaries, and an instruction-conditioned full-video response.

\subsection{Streaming QA Synthesis}

Existing streaming QA datasets often provide poorly calibrated response-time supervision.
Some response targets are assigned before sufficient visual evidence is available, which can encourage reliance on language priors and increase the risk of hallucinated answers.
Others are assigned to the end of a grounding interval even when sufficient evidence is available much earlier.
The QA branch therefore follows a simple timing principle: each response should be assigned to an earlier point, provided that the available evidence is sufficient to support the answer.

\textbf{\textit{Task-directed streaming QA design.}}
We first collect metadata from two complementary sources: verified streaming caption annotations and selected annotations from existing datasets. We then define a set of target capabilities for streaming interaction. For each capability, we prompt Gemini and Seed with a task-specific template to generate question-answer pairs grounded in the source annotations and tailored to streaming interaction.
\textbf{\textit{Coarse evidence interval localization.}}
For each synthesized question-answer pair, we provide the video, question, and reference answer to a VLM to localize a coarse temporal interval containing the visual evidence needed to support the answer. To verify the localized interval, we provide only the corresponding video clip to a VLM and ask it to answer the question. We discard examples for which the model fails to produce correct answer.
\textbf{\textit{Fine-grained response-time calibration.}}
Within each coarse interval, we evaluate candidate timestamps using sliding windows. We score reasoning-oriented QA by the conditional likelihood of the reference answer given each video prefix, and perception-oriented tasks by the visual-text similarity between each window and the target event. We select the earliest timestamp whose score satisfies the corresponding reliability criterion as the calibrated response time. Together with the localized evidence interval, this timestamp defines the streaming QA sequence: silence before relevant evidence appears, standby as supporting evidence emerges, and response at the calibrated time.

\section{Experiments}

\textbf{Implementation Details.}
We initialize OneStreamer from Qwen3-VL-4B-Instruct and perform single-stage supervised fine-tuning on the OneStreamer-1M streaming video interaction dataset together with offline data sampled from LLaVA-Video~\cite{zhang2024llavavideo}. We freeze the vision encoder and optimize the multimodal projector and LLM for one epoch, with a maximum sequence length of 131,072 tokens and a peak learning rate of $1\times10^{-5}$. Training is conducted on 32 NVIDIA H200 GPUs. Full training and data-processing configurations are provided in Appendix~\ref{sec:app:implementation}.

\subsection{Main Results}

Across the four perception and memory benchmarks in Table~\ref{tab:online_benchmarks}, {\ours} achieves the highest aggregate scores among the compared methods: 72.1 on OVOBench~\cite{niu2025ovobench}, 86.9 on StreamingBench Real-Time~\cite{lin2026streamingbench}, 66.8 on OVBench~\cite{huang2025videochatonline}, and 71.3 on ODVBench~\cite{zeng2026streamforest}. At 4B parameters, it outperforms the size-matched Qwen3-VL base model by 13.3, 5.1, 11.4, and 13.7 points, respectively. It also surpasses the 11B MOSS-VL-Realtime by 1.9, 4.0, 13.1, and 7.4 points on the same benchmarks. Overall, these results demonstrate strong performance in both real-time perception and long-range online understanding.

\begin{table}[t]
    \centering
    \renewcommand{\arraystretch}{1.2}
    \setlength{\tabcolsep}{2pt}
    \caption{
        \textbf{Online video benchmark results.} The four benchmarks on the left evaluate perception and memory, while the four on the right evaluate proactive response. With only 4B parameters, OneStreamer achieves the highest score among the compared methods on all eight benchmarks, showing consistent gains across online perception, memory, and proactive response.
    }
    \label{tab:online_benchmarks}

    \resizebox{1\textwidth}{!}{%
        \begin{tabular}{lc|cccc|cccc}
            \hline \hline
            \multirow{3}{*}{\textbf{Method}} & \multirow{3}{*}{\textbf{Size}} & \multicolumn{4}{c|}{\textbf{Perception \& Memory}} & \multicolumn{4}{c}{\textbf{Proactive Response}} \\ \cline{3-6} \cline{7-10}
             & & \textbf{OVOBench} & \textbf{StreamingBench} & \textbf{OVBench} & \textbf{ODVBench} & \textbf{ProactiveVQA} & \textbf{OmniMMI} & \textbf{OVO-Timing} & \textbf{ViSpeak} \\
             & & Overall & Real-Time & Avg. & Overall & Avg. & Avg. & Avg. F1 & Avg. \\
            \hline
            \multicolumn{10}{l}{\textit{Online video methods}} \\
            \hline
            VideoLLM-Online~\cite{chen2024videollmonline} & 7B & 12.8 & 36.0 & 9.6 & - & 23.6 & - & 6.9 & - \\
            Flash-Vstream~\cite{zhang2024flashvstream} & 7B & 33.2 & 23.2 & 31.2 & 35.7 & - & - & - & - \\
            Dispider~\cite{qian2025dispider} & 7B+1.5B & 41.8 & 67.6 & - & 45.2 & - & - & - & - \\
            VideoChat-Online~\cite{huang2025videochatonline} & 4B & - & - & 54.9 & 54.5 & - & - & - & - \\
            TimeChat-Online~\cite{yao2025timechatonline} & 7B & 47.6 & 75.4 & - & - & - & - & - & - \\
            StreamBridge~\cite{wang2026streambridge} & 7B+0.5B & 62.6 & 77.0 & - & - & - & - & - & - \\
            StreamForest~\cite{zeng2026streamforest} & 7B & 55.6 & 77.3 & 65.4 & 59.9 & - & - & - & - \\
            StreamingVLM~\cite{xu2026streamingvlm} & 7B & - & - & - & - & 17.9 & - & - & - \\
            MMDuet-2~\cite{wang2026mmduet2} & 3B & - & - & - & - & 39.8 & - & 20.5 & - \\
            Streamo~\cite{xia2026streamo} & 7B & 57.9 & - & - & - & - & - & - & - \\
            Em-Garde~\cite{zheng2026emgarde} & 7B+2B & - & - & - & - & - & - & 31.0 & - \\
            VideoChat3~\cite{li2026videochat3} & 4B & 58.5 & 81.9 & 62.5 & 70.8 & 37.6 & 24.6 & 33.6 & 1.05 \\
            Mage-VL~\cite{yang2026magevl} & 4B & 58.5 & 82.1 & 57.5 & 64.1 & 25.6 & 15.6 & 21.7 & 0.85 \\
            JoyAI-VL-Interaction~\cite{yao2026joyaivlinteraction} & 8B & 59.1 & 82.7 & 62.3 & 68.6 & 29.6 & 17.8 & 20.0 & 2.16 \\
            AURA~\cite{lu2026aura} & 8B & 65.3 & 83.2 & 58.3 & 58.8 & 30.8 & 25.4 & 11.1 & 0.87 \\
            MOSS-VL-Realtime~\cite{wang2026mossvl} & 11B & 70.2 & 82.9 & 53.7 & 63.9 & 47.2 & 32.7 & 38.5 & 2.48 \\
            Qwen3-VL (base)~\cite{bai2025qwen3vl} & 4B & 58.8 & 81.8 & 55.4 & 57.6 & 34.3 & 29.4 & 29.4 & 2.41 \\
            \hline
            \rowcolor[HTML]{DCEEFC}
            \textbf{OneStreamer} & \textbf{4B} & \textbf{72.1} & \textbf{86.9} & \textbf{66.8} & \textbf{71.3} & \textbf{48.7} & \textbf{36.6} & \textbf{41.6} & \textbf{2.87} \\
            \hline \hline
        \end{tabular}%
    }
    
\end{table}

Across the four proactive-response benchmarks in Table~\ref{tab:online_benchmarks}, {\ours} achieves the highest aggregate scores among the compared methods: 48.7 on ProactiveVQA~\cite{wang2025proactivevideoqa}, 36.6 on OmniMMI~\cite{wang2025omnimmi} with ASR, 41.6 on OVO-Timing~\cite{zheng2026emgarde}, and 2.87 on ViSpeak~\cite{fu2025vispeak}. On the first three benchmarks, it outperforms the size-matched Qwen3-VL base model by 14.4, 7.2, and 12.2 points, respectively. It also surpasses the 11B MOSS-VL-Realtime by 1.5, 3.9, and 3.1 points on the same benchmarks. On ViSpeak, its overall score reaches 2.87, compared with 2.41 for Qwen3-VL and 2.48 for MOSS-VL-Realtime. Overall, these results demonstrate consistent gains across diverse proactive-response settings.

\subsection{Ablations and Findings}

\textbf{Finding 1: Proactively generated captions provide reusable temporal memory beyond the recent visual window.}
Table~\ref{tab:ablation_caption_memory} compares FIFO and PHCM using the same checkpoint and Recent-16 visual window. The two settings differ only in whether generated caption records are retained for subsequent predictions. On OVOBench-Backward, retaining these records improves ASI from 63.5 to 71.6 and EPM from 62.0 to 62.6. These results support proactive caption generation as query-independent memory writing: records produced during observation remain useful for later QA after their source frames leave the visual window.

\begin{table}[ht]
    \centering
    \renewcommand{\arraystretch}{1.2}
    \setlength{\tabcolsep}{6pt}

    \caption{
        \textbf{Effects of Proactive Hierarchical Caption Memory.}
        Full retains the complete visual history, FIFO keeps only the recent visual window, and PHCM augments the same recent visual window with hierarchical caption memory of earlier observations.
    }
    \resizebox{1\textwidth}{!}{
        \begin{tabular}{lcc|cccc|c}
            \hline
            \multirow{2}{*}{\textbf{Variant}} & \multirow{2}{*}{\makecell{\textbf{Visual} \\ \textbf{Context}}} & \multirow{2}{*}{\makecell{\textbf{Caption} \\ \textbf{Memory}}} & \multicolumn{4}{c|}{\textbf{OVOBench}} & \multicolumn{1}{c}{\textbf{StreamingBench}} \\ \cline{4-8}
             & & & \textbf{Backward (ASI)} & \textbf{Backward (EPM)} & \textbf{Real-Time (Avg)} & \textbf{Overall} & \textbf{Real-Time (Avg)} \\ \hline

            Full & Full History & $\times$ & 67.6 & \textbf{63.0} & 70.9 & 70.9 & 80.5 \\

            FIFO & Recent-16 & $\times$ & 63.5 & 62.0 & 80.9 & 70.1 & 86.3 \\

            \rowcolor[HTML]{DCEEFC}
            \textbf{PHCM (Ours)} & Recent-16 & $\checkmark$ & \textbf{71.6} & 62.6 & \textbf{81.4} & \textbf{72.1} & \textbf{86.9} \\

            \hline
        \end{tabular}
    }

    \label{tab:ablation_caption_memory}
\end{table}

\textbf{Finding 2: Combining recent visual context with distant caption memory jointly improves perception and memory rather than merely balancing them.}
Table~\ref{tab:ablation_caption_memory} contrasts two visual-only history representations with complementary strengths. FIFO performs better on real-time perception, whereas Full scores higher on both ASI and EPM. This contrast is consistent with prior observations that extensive visual history can interfere with current perception~\citep{shen2026simplestream}. PHCM does not merely occupy an intermediate point between these baselines. Using the same Recent-16 visual window as FIFO, it improves OVOBench Real-Time from 80.9 to 81.4 and StreamingBench Real-Time from 86.3 to 86.9. The same configuration also outperforms Full on ASI by 4.0 points (71.6 vs. 67.6), despite replacing distant visual history with caption records. Its EPM score is slightly lower than Full (62.6 vs. 63.0). PHCM also achieves the highest OVOBench Overall score of 72.1, compared with 70.9 for Full and 70.1 for FIFO. Together, these results demonstrate more than a favorable compromise: PHCM exceeds FIFO's real-time perception and Full's memory performance within a single representation, rather than sacrificing one capability to improve the other.

\begin{table}[ht]
    \centering
    \renewcommand{\arraystretch}{1.2}
    \setlength{\tabcolsep}{6pt}

    \caption{
        \textbf{Ablation of state-token supervision.}
        Random Sparse randomly selects state tokens to match PSTL's supervision ratio, while Transition Only selects only state-change tokens.
        Focal applies the frequency-balanced focal loss introduced by Streamo~\citep{xia2026streamo} to all state tokens.
    }
    \resizebox{1\textwidth}{!}{
        \begin{tabular}{lcc|ccc}
            \hline
            \textbf{State Selection} & \textbf{State Loss} & \textbf{Supervision Ratio} & \textbf{ProactiveVQA (Avg)} & \textbf{OmniMMI (Avg)} & \textbf{OVO-Timing (Avg F1)} \\ \hline

            All state tokens & CE & 100.0 \% & 26.1 & 30.8 & 1.5 \\

            All state tokens & Focal & 100.0 \% & 47.0 & 32.8 & 26.5 \\

            Random Sparse & CE & 27.5 \% & 26.6 & 27.4 & 1.4 \\

            Transition Only & CE & \textbf{18.1 \%} & 46.7 & 27.4 & 15.3 \\

            \rowcolor[HTML]{DCEEFC}
            \textbf{PSTL (Ours)} & CE & 27.5 \% & \textbf{48.7} & \textbf{36.6} & \textbf{41.6} \\

            \hline
        \end{tabular}
    }

    \label{tab:ablation_state_supervision}
\end{table}

\textbf{Finding 3: Selective supervision of state transitions and persistence outperforms dense state-token supervision for proactive response learning.}
Table~\ref{tab:ablation_state_supervision} compares state-token supervision strategies under the same training data, input sequences, and text supervision. Under dense supervision, frequency-balanced focal loss substantially outperforms cross-entropy, consistent with the adverse effects of state-label imbalance. Transition Only remains competitive on ProactiveVQA but trails PSTL on OmniMMI and OVO-Timing, suggesting the value of supervising state persistence alongside transitions. PSTL preserves supervision at all output anchors and selects representative state-change and state-persistence tokens. With only 27.5\% of annotated state tokens supervised, it achieves the best results on all three benchmarks: 48.7 on ProactiveVQA, 36.6 on OmniMMI, and 41.6 on OVO-Timing. Random Sparse performs substantially worse at the same supervision ratio, indicating that sparsity alone does not explain PSTL's gains. Overall, PSTL improves proactive-response performance over dense supervision while supervising fewer state tokens.

\subsection{Efficiency and Latency}

\begin{table}[t]
    \centering
    \renewcommand{\arraystretch}{1.2}
    \setlength{\tabcolsep}{4pt}

    \caption{
        \textbf{Answer-stage efficiency and latency.}
        Full retains the complete visual history, while FIFO and PHCM share the same recent visual window. PHCM additionally uses precomputed caption memory.
        Measurements are obtained from a 360\,s OVOBench sample on a single NVIDIA H200.
    }

    \resizebox{1\textwidth}{!}{
        \begin{tabular}{lcc|ccc}
            \hline
            \textbf{Strategy} & \textbf{Visual Context} & \textbf{Caption Memory} & \textbf{GPU Memory (GB)} $\downarrow$ & \textbf{Context Tokens} $\downarrow$ & \textbf{TTFT (s)} $\downarrow$ \\ \hline

            Full & Full History & $\times$ & 25.18 & 62,094 & 4.560 \\

            FIFO & Recent-16 & $\times$ & 9.69 & 3,036 & 0.094 \\

            \rowcolor[HTML]{DCEEFC}
            PHCM (Ours) & Recent-16 & $\checkmark$ & 9.98 & 4,308 & 0.124 \\

            \hline
        \end{tabular}
    }

    \label{tab:efficiency_latency}
\end{table}

Table~\ref{tab:efficiency_latency} compares the answer-stage efficiency of Full, FIFO, and PHCM on a 360\,s OVOBench sample. Retaining the full visual history incurs the highest cost, requiring 62,094 context tokens and 25.18\,GB of GPU memory, with a TTFT of 4.560\,s. PHCM reduces the context length by 93.1\% and GPU memory by 60.4\% relative to Full, while reducing TTFT from 4.560\,s to 0.124\,s. Compared with FIFO, PHCM adds only 1,272 context tokens, 0.29\,GB of GPU memory, and 0.030\,s of TTFT. Together with the memory and perception gains in Table~\ref{tab:ablation_caption_memory}, this comparison shows that PHCM extends access to distant evidence with modest answer-stage overhead over FIFO, while remaining substantially more efficient than retaining the full visual history.

\subsection{Qualitative Analysis}

\begin{figure*}[t]
    \centering
    \includegraphics[width=\textwidth]{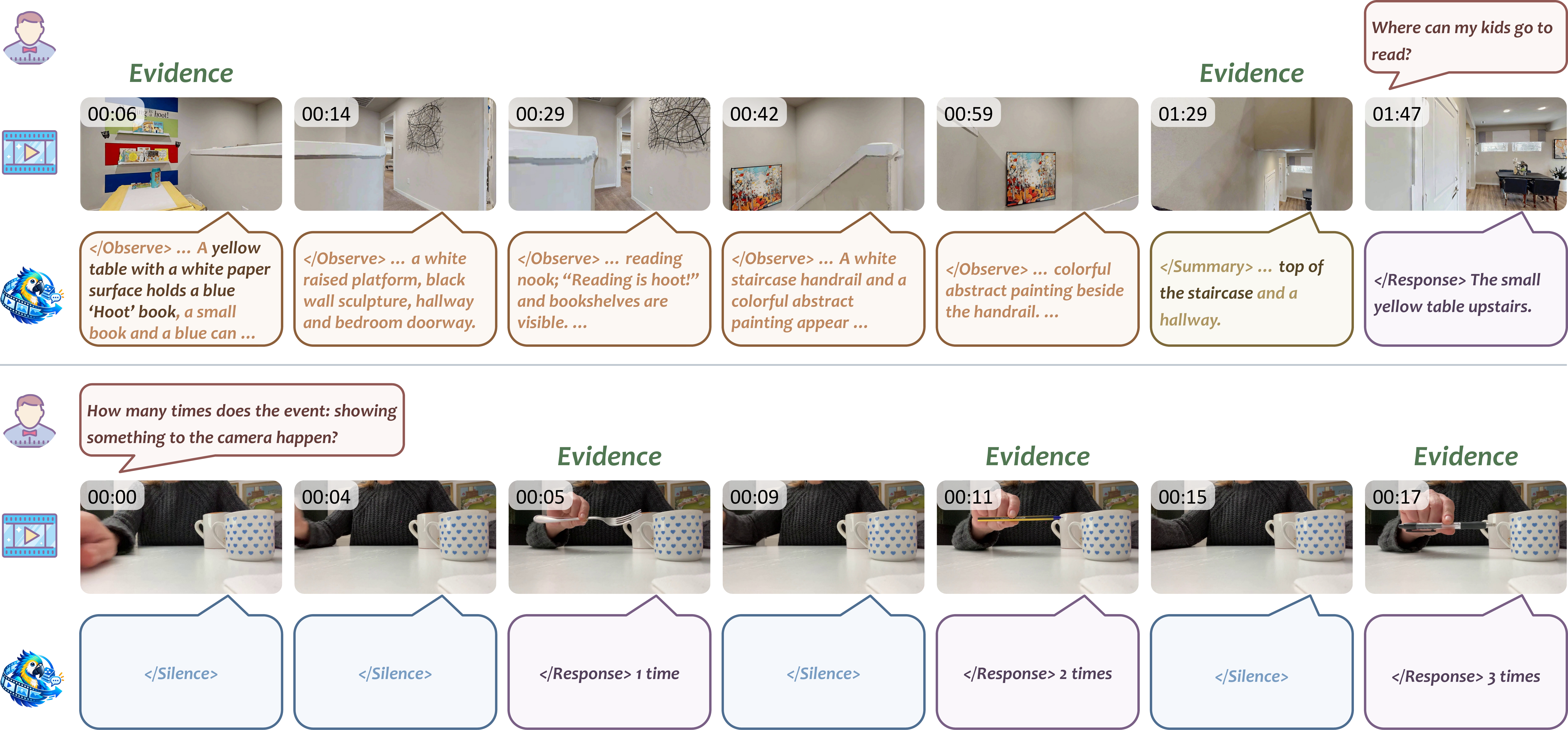}
    \caption{\textbf{Qualitative examples of memory and proactive response.} PHCM preserves past evidence beyond the recent visual window. The model remains silent until sufficient evidence arrives.}
    \label{fig:qualitative_analysis}
\end{figure*}

Figure~\ref{fig:qualitative_analysis} illustrates how {\ours} retains past evidence for later use and produces responses as relevant evidence becomes available. In the upper demo, PHCM records local details about the yellow table and books with \texttt{</Observe>} and retains higher-level scene context with \texttt{</Summary>}. When the user later asks where the children can go to read, the relevant frames have already left the recent visual window. The retained records nevertheless preserve the evidence needed to answer, leading to \emph{``The small yellow table upstairs.''} In the lower demo, the model performs proactive counting. It remains in \texttt{</Silence>} between target events and emits \texttt{</Response>} whenever new evidence supports an updated count, progressing from one to three. Together, these examples show how PHCM preserves useful evidence beyond the recent visual context, while response-control states enable timely outputs as new evidence becomes available.

\section{Conclusion}

We present {\ours}, a streaming video LLM that jointly learns query-independent evidence recording and task response through proactive generation.
PHCM turns observed content into time-grounded local-detail captions and summaries of completed events that complement a recent visual window.
PSTL preserves supervision at all output anchors and selects representative state-change and state-persistence tokens to reduce the dominance of repeated waiting states.
Our streaming data synthesis pipeline aligns caption and QA targets with available evidence in both content and timing, supporting the construction of OneStreamer-1M.
The resulting 4B model achieves the best results among the compared methods across all eight streaming video understanding benchmarks.
Ablations show that retaining generated captions improves historical QA without degrading real-time perception.
PSTL also outperforms dense state supervision and a supervision-matched random baseline, indicating that its gains are not due to sparsity alone.
Together, these findings support proactive generation as a shared learning mechanism for building reusable factual memory and producing timely responses grounded in past and current evidence.


\bibliographystyle{plainnat}
\bibliography{references}

\begin{thebibliography}{47}
\providecommand{\natexlab}[1]{#1}
\providecommand{\url}[1]{\texttt{#1}}
\expandafter\ifx\csname urlstyle\endcsname\relax
  \providecommand{\doi}[1]{doi: #1}\else
  \providecommand{\doi}{doi: \begingroup \urlstyle{rm}\Url}\fi

\bibitem[Azad et~al.(2026)Azad, Vineet, and Rawat]{azad2026streamready}
Shehreen Azad, Vibhav Vineet, and Yogesh~S Rawat.
\newblock Streamready: Learning what to answer and when in long streaming
  videos.
\newblock In \emph{Proceedings of the IEEE/CVF Conference on Computer Vision
  and Pattern Recognition}, pages 40494--40504, 2026.

\bibitem[Bai et~al.(2025)Bai, Cai, Chen, Chen, Chen, Cheng, Deng, Ding, Gao,
  Ge, et~al.]{bai2025qwen3vl}
Shuai Bai, Yuxuan Cai, Ruizhe Chen, Keqin Chen, Xionghui Chen, Zesen Cheng,
  Lianghao Deng, Wei Ding, Chang Gao, Chunjiang Ge, et~al.
\newblock Qwen3-vl technical report.
\newblock \emph{arXiv preprint arXiv:2511.21631}, 2025.

\bibitem[Chen et~al.(2024)Chen, Lv, Wu, Lin, Song, Gao, Liu, Gao, Mao, and
  Shou]{chen2024videollmonline}
Joya Chen, Zhaoyang Lv, Shiwei Wu, Kevin~Qinghong Lin, Chenan Song, Difei Gao,
  Jia-Wei Liu, Ziteng Gao, Dongxing Mao, and Mike~Zheng Shou.
\newblock Videollm-online: Online video large language model for streaming
  video.
\newblock In \emph{2024 IEEE/CVF Conference on Computer Vision and Pattern
  Recognition (CVPR)}, pages 18407--18418. IEEE, 2024.

\bibitem[Chen et~al.(2025)Chen, Zeng, Lin, Li, Ma, and Shou]{chen2025livecc}
Joya Chen, Ziyun Zeng, Yiqi Lin, Wei Li, Zejun Ma, and Mike~Zheng Shou.
\newblock Livecc: Learning video llm with streaming speech transcription at
  scale.
\newblock In \emph{2025 IEEE/CVF Conference on Computer Vision and Pattern
  Recognition (CVPR)}, pages 29083--29095. IEEE, 2025.

\bibitem[Ding et~al.(2025)Ding, Wu, Yang, Jiang, Zhang, Bai, Chen, and
  Cao]{ding2025streammind}
Xin Ding, Hao Wu, Yifan Yang, Shiqi Jiang, Qianxi Zhang, Donglin Bai, Zhibo
  Chen, and Ting Cao.
\newblock Streammind: Unlocking full frame rate streaming video dialogue
  through event-gated cognition.
\newblock In \emph{2025 IEEE/CVF International Conference on Computer Vision
  (ICCV)}, pages 13448--13459. IEEE, 2025.

\bibitem[Fu et~al.(2025)Fu, Yang, Li, Peng, Lin, Wei, Hu, Xie, and
  Zheng]{fu2025vispeak}
Shenghao Fu, Qize Yang, Yuan-Ming Li, Yi-Xing Peng, Kun-Yu Lin, Xihan Wei,
  Jian-Fang Hu, Xiaohua Xie, and Wei-Shi Zheng.
\newblock Vispeak: Visual instruction feedback in streaming videos.
\newblock In \emph{2025 IEEE/CVF International Conference on Computer Vision
  (ICCV)}, pages 21778--21788. IEEE, 2025.

\bibitem[Guan et~al.(2026)Guan, Yin, Liang, Ju, Luo, Luan, Liu, and
  Bai]{guan2026videostreamingthinking}
Yiran Guan, Liang Yin, Dingkang Liang, Jianzhong Ju, Zhenbo Luo, Jian Luan,
  Yuliang Liu, and Xiang Bai.
\newblock Video streaming thinking: Videollms can watch and think
  simultaneously.
\newblock In \emph{European Conference on Computer Vision}, pages 80--99.
  Springer, 2026.

\bibitem[Huang et~al.(2025)Huang, Li, Li, Wang, Zeng, Liang, Wu, Chen, Li, and
  Wang]{huang2025videochatonline}
Zhenpeng Huang, Xinhao Li, Jiaqi Li, Jing Wang, Xiangyu Zeng, Cheng Liang, Tao
  Wu, Xi~Chen, Liang Li, and Limin Wang.
\newblock Online video understanding: Ovbench and videochat-online.
\newblock In \emph{2025 IEEE/CVF Conference on Computer Vision and Pattern
  Recognition (CVPR)}, pages 3328--3338. IEEE, 2025.

\bibitem[Li et~al.(2025)Li, Hu, Shao, Shen, and Nie]{li2025lionfs}
Wei Li, Bing Hu, Rui Shao, Leyang Shen, and Liqiang Nie.
\newblock Lion-fs: Fast \& slow video-language thinker as online video
  assistant.
\newblock In \emph{2025 IEEE/CVF Conference on Computer Vision and Pattern
  Recognition (CVPR)}, pages 3240--3251. IEEE, 2025.

\bibitem[Li et~al.(2026{\natexlab{a}})Li, Wang, Yu, Zeng, Zhu, Huang, Gao, Li,
  He, Wang, et~al.]{li2026videochatflash}
Xinhao Li, Yi~Wang, Jiashuo Yu, Xiangyu Zeng, Yuhan Zhu, Haian Huang, Jianfei
  Gao, Kunchang Li, Yinan He, Chenting Wang, et~al.
\newblock Videochat-flash: Hierarchical compression for long-context video
  modeling.
\newblock In \emph{International Conference on Learning Representations},
  volume 2026, pages 109089--109117, 2026{\natexlab{a}}.

\bibitem[Li et~al.(2026{\natexlab{b}})Li, Zhu, Zeng, Dong, Wu, Zhang, Yang, Ma,
  Zhang, Shi, et~al.]{li2026videochat3}
Xinhao Li, Yuhan Zhu, Xiangyu Zeng, Yuhao Dong, Haoning Wu, Zhiqiu Zhang,
  Yuandong Yang, Changlian Ma, Qingyu Zhang, Yansong Shi, et~al.
\newblock Videochat3: Fully open video mllm for efficient and generalist video
  understanding.
\newblock \emph{arXiv preprint arXiv:2607.14935}, 2026{\natexlab{b}}.

\bibitem[Lin et~al.(2026)Lin, Fang, Chen, Cheng, Wan, Luo, Wang, Li, Liu, and
  Sun]{lin2026streamingbench}
Junming Lin, Zheng Fang, Chi Chen, Haoxuan Cheng, Zihao Wan, Fuwen Luo, Ziyue
  Wang, Peng Li, Yang Liu, and Maosong Sun.
\newblock Streamingbench: Assessing the gap for mllms to achieve streaming
  video understanding.
\newblock In \emph{ICASSP 2026-2026 IEEE International Conference on Acoustics,
  Speech and Signal Processing (ICASSP)}, pages 12147--12151. IEEE, 2026.

\bibitem[Liu et~al.(2026)Liu, Guo, Li, Zhen, He, Ji, Ren, Zhang, Lu, and
  Liu]{liu2026thinkingstreamingvideo}
Zikang Liu, Longteng Guo, Handong Li, Ru~Zhen, Xingjian He, Ruyi Ji, Xiaoming
  Ren, Yanhao Zhang, Haonan Lu, and Jing Liu.
\newblock Thinking in streaming video.
\newblock \emph{arXiv preprint arXiv:2603.12938}, 2026.

\bibitem[Lu et~al.(2026)Lu, Bo, Chen, Li, Guo, Guan, Liu, Xu, Sun, Sun,
  et~al.]{lu2026aura}
Xudong Lu, Yang Bo, Jinpeng Chen, Shuhan Li, Xintong Guo, Huankang Guan, Fang
  Liu, Dunyuan Xu, Peiwen Sun, Heyang Sun, et~al.
\newblock Aura: Always-on understanding and real-time assistance via video
  streams.
\newblock \emph{arXiv preprint arXiv:2604.04184}, 2026.

\bibitem[Niu et~al.(2025)Niu, Li, Miao, Ge, Zhou, He, Dong, Duan, Ding, Qian,
  et~al.]{niu2025ovobench}
Junbo Niu, Yifei Li, Ziyang Miao, Chunjiang Ge, Yuanhang Zhou, Qihao He, Xiaoyi
  Dong, Haodong Duan, Shuangrui Ding, Rui Qian, et~al.
\newblock Ovo-bench: How far is your video-llms from real-world online video
  understanding?
\newblock In \emph{2025 IEEE/CVF Conference on Computer Vision and Pattern
  Recognition (CVPR)}, pages 18902--18913. IEEE, 2025.

\bibitem[Qian et~al.(2025)Qian, Ding, Dong, Zhang, Zang, Cao, Lin, and
  Wang]{qian2025dispider}
Rui Qian, Shuangrui Ding, Xiaoyi Dong, Pan Zhang, Yuhang Zang, Yuhang Cao,
  Dahua Lin, and Jiaqi Wang.
\newblock Dispider: Enabling video llms with active real-time interaction via
  disentangled perception, decision, and reaction.
\newblock In \emph{2025 IEEE/CVF Conference on Computer Vision and Pattern
  Recognition (CVPR)}, pages 24045--24055. IEEE, 2025.

\bibitem[Shen et~al.(2024)Shen, Xiong, Zhao, Wu, Chen, Zhu, Liu, Xiao,
  Varadarajan, Bordes, et~al.]{shen2024longvu}
Xiaoqian Shen, Yunyang Xiong, Changsheng Zhao, Lemeng Wu, Jun Chen, Chenchen
  Zhu, Zechun Liu, Fanyi Xiao, Balakrishnan Varadarajan, Florian Bordes, et~al.
\newblock Longvu: Spatiotemporal adaptive compression for long video-language
  understanding.
\newblock \emph{arXiv preprint arXiv:2410.17434}, 2024.

\bibitem[Shen et~al.(2026)Shen, Tian, Yang, and Liu]{shen2026simplestream}
Yujiao Shen, Shulin Tian, Jingkang Yang, and Ziwei Liu.
\newblock A simple baseline for streaming video understanding.
\newblock \emph{arXiv preprint arXiv:2604.02317}, 2026.

\bibitem[Song et~al.(2024)Song, Chai, Wang, Zhang, Zhou, Wu, Chi, Guo, Ye,
  Zhang, et~al.]{song2024moviechat}
Enxin Song, Wenhao Chai, Guanhong Wang, Yucheng Zhang, Haoyang Zhou, Feiyang
  Wu, Haozhe Chi, Xun Guo, Tian Ye, Yanting Zhang, et~al.
\newblock Moviechat: From dense token to sparse memory for long video
  understanding.
\newblock In \emph{2024 IEEE/CVF Conference on Computer Vision and Pattern
  Recognition (CVPR)}, pages 18221--18232. IEEE, 2024.

\bibitem[Wang et~al.(2026{\natexlab{a}})Wang, Tan, Cao, Li, and
  Chen]{wang2026curvestream}
Chao Wang, Xudong Tan, Jianjian Cao, Kangcong Li, and Tao Chen.
\newblock Curvestream: boosting streaming video understanding in mllms via
  curvature-aware hierarchical visual memory management.
\newblock In \emph{European Conference on Computer Vision}, pages 605--623.
  Springer, 2026{\natexlab{a}}.

\bibitem[Wang et~al.(2025{\natexlab{a}})Wang, Feng, Lai, Xu, Li, Ge, Dehghan,
  Cao, and Huang]{wang2026streambridge}
Haibo Wang, Bo~Feng, Zhengfeng Lai, Mingze Xu, Shiyu Li, Weifeng Ge, Afshin
  Dehghan, Meng Cao, and Ping Huang.
\newblock Streambridge: Turning your offline video large language model into a
  proactive streaming assistant.
\newblock \emph{Advances in Neural Information Processing Systems},
  38:\penalty0 132332--132359, 2025{\natexlab{a}}.

\bibitem[Wang et~al.(2026{\natexlab{b}})Wang, Jin, Hao, Chen, Liu, Ao, and
  Zhao]{wang2026thinkwhilewatching}
Lu~Wang, Zhuoran Jin, Yupu Hao, Yubo Chen, Kang Liu, Yulong Ao, and Jun Zhao.
\newblock Think while watching: Online streaming segment-level memory for
  multi-turn video reasoning in multimodal large language models.
\newblock \emph{arXiv preprint arXiv:2603.11896}, 2026{\natexlab{b}}.

\bibitem[Wang et~al.(2026{\natexlab{c}})Wang, Tan, Zhou, Zhou, Chen, He, Zeng,
  Cheng, Wang, Qian, et~al.]{wang2026mossvl}
Pengyu Wang, Chenkun Tan, Shaojun Zhou, Qirui Zhou, Yanxin Chen, Xingyang He,
  Huazheng Zeng, Jijun Cheng, Chenghao Wang, Xiaomeng Qian, et~al.
\newblock Moss-vl technical report.
\newblock \emph{arXiv preprint arXiv:2608.15045}, 2026{\natexlab{c}}.

\bibitem[Wang et~al.(2025{\natexlab{b}})Wang, Li, Yan, He, Yu, Zeng, Wang, Ma,
  Huang, Gao, et~al.]{wang2025internvideo25}
Yi~Wang, Xinhao Li, Ziang Yan, Yinan He, Jiashuo Yu, Xiangyu Zeng, Chenting
  Wang, Changlian Ma, Haian Huang, Jianfei Gao, et~al.
\newblock Internvideo2.5: Empowering video mllms with long and rich context
  modeling.
\newblock \emph{arXiv preprint arXiv:2501.12386}, 2025{\natexlab{b}}.

\bibitem[Wang et~al.(2025{\natexlab{c}})Wang, Meng, Wang, Zhang, and
  Zhao]{wang2025proactivevideoqa}
Yueqian Wang, Xiaojun Meng, Yifan Wang, Huishuai Zhang, and Dongyan Zhao.
\newblock Proactivevideoqa: A comprehensive benchmark evaluating proactive
  interactions in video large language models.
\newblock \emph{arXiv preprint arXiv:2507.09313}, 2025{\natexlab{c}}.

\bibitem[Wang et~al.(2026{\natexlab{d}})Wang, Liu, Wang, Xu, Wan, Zhang, and
  Zhao]{wang2026mmduet2}
Yueqian Wang, Songxiang Liu, Disong Wang, Nuo Xu, Guanglu Wan, Huishuai Zhang,
  and Dongyan Zhao.
\newblock Mmduet2: Enhancing proactive interaction of video mllms with
  multi-turn reinforcement learning.
\newblock In \emph{International Conference on Learning Representations},
  volume 2026, pages 23257--23270, 2026{\natexlab{d}}.

\bibitem[Wang et~al.(2025{\natexlab{d}})Wang, Wang, Chen, Wu, Zhao, and
  Zheng]{wang2025omnimmi}
Yuxuan Wang, Yueqian Wang, Bo~Chen, Tong Wu, Dongyan Zhao, and Zilong Zheng.
\newblock Omnimmi: A comprehensive multi-modal interaction benchmark in
  streaming video contexts.
\newblock In \emph{2025 IEEE/CVF Conference on Computer Vision and Pattern
  Recognition (CVPR)}, pages 18925--18935. IEEE, 2025{\natexlab{d}}.

\bibitem[Xia et~al.(2026)Xia, Chen, Zhang, Sun, and Zhou]{xia2026streamo}
Jiaer Xia, Peixian Chen, Mengdan Zhang, Xing Sun, and Kaiyang Zhou.
\newblock Streaming video instruction tuning.
\newblock In \emph{Proceedings of the IEEE/CVF Conference on Computer Vision
  and Pattern Recognition}, pages 31219--31229, 2026.

\bibitem[Xie et~al.(2026{\natexlab{a}})Xie, Huang, Tan, Wang, Zeng, Wu, Chen,
  Wang, and Fu]{xie2026streamov}
Ming Xie, Zizheng Huang, Xudong Tan, Chao Wang, Xiangyu Zeng, Wenxiao Wu, Tao
  Chen, Limin Wang, and Yanwei Fu.
\newblock Streamov: Streaming omni-video understanding via evidence-guided
  memory and response triggering.
\newblock \emph{arXiv preprint arXiv:2605.25621}, 2026{\natexlab{a}}.

\bibitem[Xie et~al.(2026{\natexlab{b}})Xie, He, Wang, Zheng, Ye, and
  Wu]{xie2026fluxmem}
Yiweng Xie, Bo~He, Junke Wang, Xiangyu Zheng, Ziyi Ye, and Zuxuan Wu.
\newblock Fluxmem: Adaptive hierarchical memory for streaming video
  understanding.
\newblock \emph{arXiv preprint arXiv:2603.02096}, 2026{\natexlab{b}}.

\bibitem[Xu et~al.(2026)Xu, Xiao, Chen, He, Peng, Lu, and
  Han]{xu2026streamingvlm}
Ruyi Xu, Guangxuan Xiao, Yukang Chen, Liuning He, Kelly Peng, Yao Lu, and Song
  Han.
\newblock Streamingvlm: Real-time understanding for infinite video streams.
\newblock In \emph{International Conference on Learning Representations},
  volume 2026, pages 61463--61475, 2026.

\bibitem[Yan et~al.(2026)Yan, Dai, Ran, Li, Lin, Jin, Xie, Liao, and
  Lian]{yan2026proactvl}
Weicai Yan, Yuhong Dai, Qi~Ran, Haodong Li, Wang Lin, Tao Jin, Xing Xie, Hao
  Liao, and Jianxun Lian.
\newblock Proact-vl: A proactive videollm for real-time ai companions.
\newblock \emph{arXiv preprint arXiv:2603.03447}, 2026.

\bibitem[Yang et~al.(2025{\natexlab{a}})Yang, Tang, Zhao, Zhuang, Lu, An, Hu,
  Zhang, Swikir, He, et~al.]{yang2025streamagent}
Haolin Yang, Feilong Tang, Lingxiao Zhao, Xinlin Zhuang, Yifan Lu, Xiang An,
  Ming Hu, Xiaofeng Zhang, Abdalla Swikir, Junjun He, et~al.
\newblock Streamagent: Towards anticipatory agents for streaming video
  understanding.
\newblock \emph{arXiv preprint arXiv:2508.01875}, 2025{\natexlab{a}}.

\bibitem[Yang et~al.(2026)Yang, Zhang, Jia, Guo, Shen, Zhang, Zhang, Wang, Li,
  Zhang, et~al.]{yang2026magevl}
Senqiao Yang, Kaichen Zhang, Zhaoyang Jia, Jinghao Guo, Yifei Shen, Xinjie
  Zhang, Xiaoyi Zhang, Haoqing Wang, Xiao Li, Peng Zhang, et~al.
\newblock Mage-vl: An efficient codec-native streaming multimodal foundation
  model.
\newblock \emph{arXiv preprint arXiv:2607.24904}, 2026.

\bibitem[Yang et~al.(2025{\natexlab{b}})Yang, Zhang, Hu, Wang, Qian, Wen, Yang,
  Gao, Dong, and Xu]{yang2026livestar}
Zhenyu Yang, Kairui Zhang, Yuhang Hu, Bing Wang, Shengsheng Qian, Bin Wen, Fan
  Yang, Tingting Gao, Weiming Dong, and Changsheng Xu.
\newblock Livestar: Live streaming assistant for real-world online video
  understanding.
\newblock \emph{Advances in Neural Information Processing Systems},
  38:\penalty0 31266--31304, 2025{\natexlab{b}}.

\bibitem[Yao et~al.(2026)Yao, Zhou, Yang, Qin, Hou, Liang, Wang, Cao, Ye, Xie,
  et~al.]{yao2026joyaivlinteraction}
Dingyu Yao, Junhao Zhou, Chenxu Yang, Chuanyu Qin, Haowen Hou, Zheming Liang,
  Congcong Wang, Yuhang Cao, Shenglong Ye, Shuai Xie, et~al.
\newblock Joyai-vl-interaction: Real-time vision-language interaction
  intelligence.
\newblock \emph{arXiv preprint arXiv:2606.14777}, 2026.

\bibitem[Yao et~al.(2025)Yao, Li, Wei, Li, Ren, Liu, Ouyang, Wang, Li, Li,
  et~al.]{yao2025timechatonline}
Linli Yao, Yicheng Li, Yuancheng Wei, Lei Li, Shuhuai Ren, Yuanxin Liu, Kun
  Ouyang, Lean Wang, Shicheng Li, Sida Li, et~al.
\newblock Timechat-online: 80\% visual tokens are naturally redundant in
  streaming videos.
\newblock In \emph{Proceedings of the 33rd ACM International Conference on
  Multimedia}, pages 10807--10816, 2025.

\bibitem[Zeng et~al.(2025{\natexlab{a}})Zeng, Li, Wang, Li, Jiang, Yan, Li,
  Shi, Yue, Wang, et~al.]{zeng2025timesuite}
Xiangyu Zeng, Kunchang Li, Chenting Wang, Xinhao Li, Tianxiang Jiang, Ziang
  Yan, Songze Li, Yansong Shi, Zhengrong Yue, Yi~Wang, et~al.
\newblock Timesuite: Improving mllms for long video understanding via grounded
  tuning.
\newblock In \emph{International Conference on Learning Representations},
  volume 2025, pages 38057--38081, 2025{\natexlab{a}}.

\bibitem[Zeng et~al.(2025{\natexlab{b}})Zeng, Qiu, Zhang, Li, Wang, Li, Yan,
  Tian, Tian, Zhao, et~al.]{zeng2026streamforest}
Xiangyu Zeng, Kefan Qiu, Qingyu Zhang, Xinhao Li, Jing Wang, Jiaxin Li, Ziang
  Yan, Kun Tian, Meng Tian, Xinhai Zhao, et~al.
\newblock Streamforest: Efficient online video understanding with persistent
  event memory.
\newblock \emph{Advances in Neural Information Processing Systems},
  38:\penalty0 75804--75835, 2025{\natexlab{b}}.

\bibitem[Zeng et~al.(2026)Zeng, Zhang, Zhu, Li, Wang, Ma, Zhang, Huang, Ouyang,
  Jiang, et~al.]{zeng2026videoo3}
Xiangyu Zeng, Zhiqiu Zhang, Yuhan Zhu, Xinhao Li, Zikang Wang, Changlian Ma,
  Qingyu Zhang, Zizheng Huang, Kun Ouyang, Tianxiang Jiang, et~al.
\newblock Video-o3: Native interleaved clue seeking for long video multi-hop
  reasoning.
\newblock \emph{arXiv preprint arXiv:2601.23224}, 2026.

\bibitem[Zhang et~al.(2024{\natexlab{a}})Zhang, Wang, Tang, Liu, Feng, Dai, and
  Jin]{zhang2024flashvstream}
Haoji Zhang, Yiqin Wang, Yansong Tang, Yong Liu, Jiashi Feng, Jifeng Dai, and
  Xiaojie Jin.
\newblock Flash-vstream: Memory-based real-time understanding for long video
  streams.
\newblock \emph{arXiv preprint arXiv:2406.08085}, 2024{\natexlab{a}}.

\bibitem[Zhang et~al.(2026)Zhang, Tong, Lin, Wu, Sun, Ma, and
  Shen]{zhang2026thinkasyousee}
Jialiang Zhang, Junlong Tong, Junyan Lin, Hao Wu, Yirong Sun, Yunpu Ma, and
  Xiaoyu Shen.
\newblock Think-as-you-see: Streaming chain-of-thought reasoning for large
  vision-language models.
\newblock \emph{arXiv preprint arXiv:2603.02872}, 2026.

\bibitem[Zhang et~al.(2024{\natexlab{b}})Zhang, Zhang, Li, Zeng, Yang, Zhang,
  Wang, Tan, Li, and Liu]{zhang2024longva}
Peiyuan Zhang, Kaichen Zhang, Bo~Li, Guangtao Zeng, Jingkang Yang, Yuanhan
  Zhang, Ziyue Wang, Haoran Tan, Chunyuan Li, and Ziwei Liu.
\newblock Long context transfer from language to vision.
\newblock \emph{arXiv preprint arXiv:2406.16852}, 2024{\natexlab{b}}.

\bibitem[Zhang et~al.(2024{\natexlab{c}})Zhang, Wu, Li, Li, Ma, Liu, and
  Li]{zhang2024llavavideo}
Yuanhan Zhang, Jinming Wu, Wei Li, Bo~Li, Zejun Ma, Ziwei Liu, and Chunyuan Li.
\newblock Llava-video: Video instruction tuning with synthetic data.
\newblock \emph{arXiv preprint arXiv:2410.02713}, 2024{\natexlab{c}}.

\bibitem[Zheng et~al.(2026)Zheng, Ding, Yang, Jiang, Wu, Zhang, Wang, Cao, and
  Liu]{zheng2026emgarde}
Yikai Zheng, Xin Ding, Yifan Yang, Shiqi Jiang, Hao Wu, Qianxi Zhang, Weijun
  Wang, Ting Cao, and Yunxin Liu.
\newblock Em-garde: A propose-match framework for proactive streaming video
  understanding.
\newblock \emph{arXiv preprint arXiv:2603.19054}, 2026.

\bibitem[Zhou et~al.(2026)Zhou, Mei, Li, Wang, Rao, and Lyu]{zhou2026wemm}
Junjie Zhou, Ke~Mei, Lei Li, Tianyi Wang, Fengyun Rao, and Jing Lyu.
\newblock Wemm-embedding: Wechat multi-modal embedding technical report.
\newblock \emph{arXiv preprint arXiv:2608.24053}, 2026.

\bibitem[Zhou et~al.(2024)Zhou, Arnab, Buch, Yan, Myers, Xiong, Nagrani, and
  Schmid]{zhou2024streamingdvc}
Xingyi Zhou, Anurag Arnab, Shyamal Buch, Shen Yan, Austin Myers, Xuehan Xiong,
  Arsha Nagrani, and Cordelia Schmid.
\newblock Streaming dense video captioning.
\newblock In \emph{2024 IEEE/CVF Conference on Computer Vision and Pattern
  Recognition (CVPR)}, pages 18243--18252. IEEE, 2024.

\end{thebibliography}

\clearpage
\appendix
\etocdepthtag.toc{appendix}
\begingroup
    \hypersetup{linktoc=all,hidelinks}
    \setlength{\parskip}{0pt}
    \etocsettagdepth{main}{none}
    \etocsettagdepth{appendix}{subsection}
    \etocsetnexttocdepth{subsection}
    \etocsettocstyle{%
        \section*{\Large Appendix Contents}
        \pdfbookmark[0]{Appendix Contents}{appendix.contents}
        \large
    }{}
    \tableofcontents
    \clearpage
\endgroup

\section{More Details of Method}

\subsection{Proactive State Transition Learning}
\label{sec:app:pstl}

PSTL is applied to task-specific control-state sequences. Table~\ref{tab:app:pstl_states} summarizes the control-state vocabulary and output-anchor set $\mathcal{A}_{\tau}$ for each training task family. For PHCM, \texttt{</Observe>} and \texttt{</Summary>} are the output anchors for local-detail captions and event summaries. For proactive QA and proactive interaction, \texttt{</Response>} serves as the output anchor for answers and task outputs.

\begin{table}[H]
    \centering
    \caption{\textbf{Task-specific control-state tokens and output-anchor sets for PSTL.}}
    \label{tab:app:pstl_states}
    \renewcommand{\arraystretch}{1.2}
    \setlength{\tabcolsep}{10pt}
    \resizebox{\textwidth}{!}{%
    \begin{tabular}{lll}
        \toprule
        \textbf{Task} & \textbf{State Control Token} & \textbf{Output Anchors} \\
        \midrule
        PHCM & \texttt{</Silence>}, \texttt{</Observe>}, \texttt{</Summary>} & \texttt{</Observe>}, \texttt{</Summary>} \\
        Proactive QA & \texttt{</Silence>}, \texttt{</Standby>}, \texttt{</Response>} & \texttt{</Response>} \\
        Proactive Interaction & \texttt{</Silence>}, \texttt{</Response>} & \texttt{</Response>} \\
        \bottomrule
    \end{tabular}
    }
\end{table}

For a given source state, PSTL retains supervision for both state changes and state persistence. For example, \texttt{</Silence>} can persist when the available evidence is insufficient and transition to an output state once sufficient evidence becomes available. State persistence does not necessarily indicate redundant waiting because consecutive \texttt{</Observe>} tokens can initiate distinct local-detail captions. Because the quota is computed once per training sequence from all source-to-anchor transitions and then shared across groups, it also caps groups whose source state has no direct transition to an output anchor.

Let $\mathcal{I}=\bigcup_{X,Y}\mathcal{I}_{X\rightarrow Y}$ denote the selected state-token indices. The state loss is
\begin{equation}
    \mathcal{L}_{\mathrm{state}} =
    -\frac{1}{|\mathcal{I}|}
    \sum_{i \in \mathcal{I}}
    \log p_{\theta}(z_i \mid c_i),
    \label{eq:state_loss}
\end{equation}
where $z_i$ denotes the annotated control token at index $i$, and $c_i$ denotes the causal visual-text context available before predicting $z_i$. Unselected control tokens remain in the teacher-forced sequence and can condition later predictions. They simply do not contribute to $\mathcal{L}_{\mathrm{state}}$ at their own indices. This masking therefore neither blocks attention nor removes tokens from the sequence. Caption text following \texttt{</Observe>} or \texttt{</Summary>} retains full next-token supervision. The same applies to answers and other task outputs following \texttt{</Response>}. At inference, PSTL is no longer applied. The model generates control tokens and text directly from its learned conditional distributions without an external transition policy.

\subsection{Online Inference with PHCM}

At inference time, OneStreamer processes the video incrementally and maintains two complementary forms of history: a Recent-\(N\) FIFO visual window and an accumulated textual caption memory. At each decision step, the current visual tokens are concatenated with the retained dialogue and previously generated caption records. When the model predicts </Observe> or </Summary>, it autoregressively generates the corresponding record and appends it to the textual history. The source visual frames may subsequently leave the FIFO window, while the generated record remains available to later predictions. No external retriever, memory encoder, or post-hoc access to discarded visual features is used. For proactive QA, the same history is provided together with the user instruction, and the model predicts </Silence>, </Standby>, or </Response> from the currently available context.

\section{More Details of Dataset}
\label{sec:app:dataset}

\subsection{Composition and Statistics}

\begin{table}[!ht]
\centering
\setlength{\tabcolsep}{4pt}
\renewcommand{\arraystretch}{1.00}
\caption{\textbf{Online task types and data sources for OneStreamer-1M.}}
\label{tab:onestreamer_data_sources}
\resizebox{\textwidth}{!}{%
\small
\begin{tabular}{>{\raggedright\arraybackslash}p{0.23\textwidth}|>{\raggedright\arraybackslash}p{0.25\textwidth}|>{\raggedright\arraybackslash}p{0.50\textwidth}>{\raggedleft\arraybackslash}p{\dimexpr0.22\textwidth-8\tabcolsep-2\arrayrulewidth\relax}}
\toprule
\rowcolor[HTML]{C0C0C0}
\textbf{Collection} & \textbf{Task} & \textbf{Source} & \textbf{Records} \\
\midrule
\rowcolor[HTML]{DCEEFC}
\multicolumn{4}{l}{\textbf{Proactive Caption Memory}}\\*
\multirow[t]{2}{=}{\raggedright OneStreamer-Core} & Hierarchical captioning & YouTube-Collected & 52,500 \\*[1pt]
\arrayrulecolor[HTML]{AEB8C4}\cline{2-4}\arrayrulecolor{black}
 & Flat captioning & YouTube-Collected & 7,500 \\[1pt]
\midrule
\rowcolor[HTML]{DCEEFC}
\multicolumn{4}{l}{\textbf{Proactive Interaction}}\\*
\multirow[t]{2}{=}{\raggedright OneStreamer-Core} & Continuous description & COIN & 799 \\*[1pt]
\arrayrulecolor[HTML]{AEB8C4}\cline{2-4}\arrayrulecolor{black}
 & Active counting & PerceptionTest, THUMOS & 7,142 \\[1pt]
\arrayrulecolor[HTML]{AEB8C4}\cline{1-4}\arrayrulecolor{black}
\multirow[t]{2}{=}{\raggedright Streamo~\cite{xia2026streamo}} & Alerts and reminders & ActivityNet, LLaVA-Video, QVHighlights, DiDeMo, QuerYD, TACoS & 68,649 \\*[1pt]
\arrayrulecolor[HTML]{AEB8C4}\cline{2-4}\arrayrulecolor{black}
 & Continuous description & COIN, YouCook2, HowToStep, ActivityNet, QVHighlights, LLaVA-Video & 58,218 \\[1pt]
\arrayrulecolor[HTML]{AEB8C4}\cline{1-4}\arrayrulecolor{black}
\multirow[t]{2}{=}{\raggedright ViSpeak~\cite{fu2025vispeak}} & Visual interaction & ViSpeak-Instruct & 9,048 \\*[1pt]
\arrayrulecolor[HTML]{AEB8C4}\cline{2-4}\arrayrulecolor{black}
 & Contextual feedback & OOPS, FunQA & 4,642 \\[1pt]
\arrayrulecolor[HTML]{AEB8C4}\cline{1-4}\arrayrulecolor{black}
\multirow[t]{4}{=}{\raggedright JoyAI-VL-Interaction~\cite{yao2026joyaivlinteraction}} & Alerts and reminders & YouTube-Collected, MM-understanding-processed, Remind-once, TimeLens-100K, Molmo2-VideoCapQA & 105,281 \\*[1pt]
\arrayrulecolor[HTML]{AEB8C4}\cline{2-4}\arrayrulecolor{black}
 & Continuous description & Shot2Story, HoloAssist & 1,003 \\*[1pt]
\arrayrulecolor[HTML]{AEB8C4}\cline{2-4}\arrayrulecolor{black}
 & Active counting & Molmo2-VideoCapQA, VideoGPT+ training data, Molmo2-AskModelAnything & 491 \\*[1pt]
\arrayrulecolor[HTML]{AEB8C4}\cline{2-4}\arrayrulecolor{black}
 & Procedural assistance & HoloAssist & 97 \\[1pt]
\arrayrulecolor[HTML]{AEB8C4}\cline{1-4}\arrayrulecolor{black}
MMDuet2~\cite{wang2026mmduet2} & Continuous description & Live-WhisperX-526K & 16,200 \\[1pt]
\midrule
\rowcolor[HTML]{DCEEFC}
\multicolumn{4}{l}{\textbf{Proactive QA}}\\*
\multirow[t]{3}{=}{\raggedright OneStreamer-Core} & General video QA & MAGQA & 36,834 \\*[1pt]
\arrayrulecolor[HTML]{AEB8C4}\cline{2-4}\arrayrulecolor{black}
 & Video-dialogue QA & TVQA & 10,000 \\*[1pt]
\arrayrulecolor[HTML]{AEB8C4}\cline{2-4}\arrayrulecolor{black}
 & Anomaly QA & UCF-Crime & 849 \\[1pt]
\arrayrulecolor[HTML]{AEB8C4}\cline{1-4}\arrayrulecolor{black}
\multirow[t]{2}{=}{\raggedright JoyAI-VL-Interaction~\cite{yao2026joyaivlinteraction}} & Temporal QA & VideoGPT+ training data & 1,824 \\*[1pt]
\arrayrulecolor[HTML]{AEB8C4}\cline{2-4}\arrayrulecolor{black}
 & Gesture and expression QA & Biaoqing-zh, Shoushi-zh & 20,782 \\[1pt]
\arrayrulecolor[HTML]{AEB8C4}\cline{1-4}\arrayrulecolor{black}
Streamo~\cite{xia2026streamo} & General video QA & Ego-TimeQA, LLaVA-Video & 90,181 \\[1pt]
\arrayrulecolor[HTML]{AEB8C4}\cline{1-4}\arrayrulecolor{black}
Seeker~\cite{zeng2026videoo3} & General video QA & LLaVA-Video, LongVideoDB, YouTube-Collected & 19,625 \\[1pt]
\arrayrulecolor[HTML]{AEB8C4}\cline{1-4}\arrayrulecolor{black}
\multirow[t]{2}{=}{\raggedright MMDuet2~\cite{wang2026mmduet2}} & General video QA & Live-WhisperX-526K & 16,226 \\*[1pt]
\arrayrulecolor[HTML]{AEB8C4}\cline{2-4}\arrayrulecolor{black}
 & Procedural QA & EgoExoLearn & 1,263 \\[1pt]
\midrule
\rowcolor[HTML]{DCEEFC}
\multicolumn{4}{l}{\textbf{Perception\&Memory QA}}\\*
\multirow[t]{7}{=}{\raggedright OneStreamer-Core} & General video QA & EgoQA, Ego4D, COIN, Shot2Story, YouTube-Collected & 84,230 \\*[1pt]
\arrayrulecolor[HTML]{AEB8C4}\cline{2-4}\arrayrulecolor{black}
 & Spatial QA & AVA, YouTube-Collected & 19,923 \\*[1pt]
\arrayrulecolor[HTML]{AEB8C4}\cline{2-4}\arrayrulecolor{black}
 & Future prediction & Ego4D (GoalStep) & 12,803 \\*[1pt]
\arrayrulecolor[HTML]{AEB8C4}\cline{2-4}\arrayrulecolor{black}
 & Procedural QA & COIN, YouTube-Collected & 12,877 \\*[1pt]
\arrayrulecolor[HTML]{AEB8C4}\cline{2-4}\arrayrulecolor{black}
 & Counting QA & PerceptionTest, THUMOS, YouTube-Collected & 22,326 \\*[1pt]
\arrayrulecolor[HTML]{AEB8C4}\cline{2-4}\arrayrulecolor{black}
 & Person identification & YouTube-Collected & 1,314 \\*[1pt]
\arrayrulecolor[HTML]{AEB8C4}\cline{2-4}\arrayrulecolor{black}
 & Answerability judgment & EgoQA, YouTube-Collected & 3,523 \\[1pt]
\arrayrulecolor[HTML]{AEB8C4}\cline{1-4}\arrayrulecolor{black}
\multirow[t]{2}{=}{\raggedright ViSpeak~\cite{fu2025vispeak}} & General video QA & ViSpeak-Instruct & 4,574 \\*[1pt]
\arrayrulecolor[HTML]{AEB8C4}\cline{2-4}\arrayrulecolor{black}
 & Gesture and expression QA & Social-IQ & 3,606 \\[1pt]
\arrayrulecolor[HTML]{AEB8C4}\cline{1-4}\arrayrulecolor{black}
\multirow[t]{4}{=}{\raggedright StreamForest~\cite{zeng2026streamforest}} & Spatial QA & LaSOT, AS-V2, RefCOCO, Visual Genome, AVA, GOT-10k, OnlineIT-Drive & 337,570 \\*[1pt]
\arrayrulecolor[HTML]{AEB8C4}\cline{2-4}\arrayrulecolor{black}
 & Temporal grounding & ActivityNet, COIN, LaSOT, Charades, HiREST, InternVid, QuerYD & 92,283 \\*[1pt]
\arrayrulecolor[HTML]{AEB8C4}\cline{2-4}\arrayrulecolor{black}
 & Future prediction & OnlineIT-Drive & 4,390 \\*[1pt]
\arrayrulecolor[HTML]{AEB8C4}\cline{2-4}\arrayrulecolor{black}
 & Risk reasoning & OnlineIT-Drive & 13,081 \\[1pt]
\arrayrulecolor[HTML]{AEB8C4}\cline{1-4}\arrayrulecolor{black}
VideoChat-OL~\cite{huang2025videochatonline} & Spatial QA & AVA & 3,110 \\[1pt]
\arrayrulecolor[HTML]{AEB8C4}\cline{1-4}\arrayrulecolor{black}
\multirow[t]{3}{=}{\raggedright VST~\cite{guan2026videostreamingthinking}} & General video QA & VST-Training-Data & 8,425 \\*[1pt]
\arrayrulecolor[HTML]{AEB8C4}\cline{2-4}\arrayrulecolor{black}
 & Counting QA & VST-Training-Data & 3,823 \\*[1pt]
\arrayrulecolor[HTML]{AEB8C4}\cline{2-4}\arrayrulecolor{black}
 & Answerability judgment & VST-Training-Data & 2,992 \\[1pt]
\midrule
\multicolumn{3}{l}{\textbf{Total listed records}} & \textbf{1,160,004} \\
\bottomrule
\end{tabular}%
}
\end{table}

Table~\ref{tab:onestreamer_data_sources} summarizes the composition of OneStreamer-1M according to its supervision role in streaming interaction. The listed data are organized into four families: Proactive Caption Memory, Proactive Interaction, Proactive QA, and Perception \& Memory QA. Proactive Caption Memory contains 60,000 hierarchical and flat captioning records designed to supervise evidence recording through local-detail captions and event-level summaries. Proactive Interaction contains 271,570 records covering continuous description, alerts and reminders, active counting, visual interaction, contextual feedback, and procedural assistance. Proactive QA contributes 197,584 question--answer records spanning general video QA, video dialogue, temporal reasoning, gesture and expression understanding, anomaly recognition, and procedural QA. The largest component is Perception \& Memory QA, with 630,850 records covering general and spatial understanding, temporal grounding, counting, future prediction, procedural reasoning, answerability judgment, person identification, and risk reasoning. Together, these four families provide complementary supervision for evidence recording, online perception and memory, and proactive task response.

The training collection combines newly constructed OneStreamer-Core data with carefully curated open-source training data for streaming video understanding.
Specifically, we incorporate training data from Streamo~\cite{xia2026streamo}, JoyAI-VL-Interaction~\cite{yao2026joyaivlinteraction}, MMDuet2~\cite{wang2026mmduet2}, ViSpeak~\cite{fu2025vispeak}, StreamForest~\cite{zeng2026streamforest}, VideoChat-OL~\cite{huang2025videochatonline}, Seeker~\cite{zeng2026videoo3}, and VST~\cite{guan2026videostreamingthinking}.

We explicitly decontaminate the training corpus against the test split of every evaluated benchmark by tracing all samples back to their original source videos. Any training sample whose source video appears in a benchmark test set is removed, including clips taken from non-overlapping temporal intervals of the same video. This source-video-level filtering prevents train--test overlap through shared video sources.

\subsection{Streaming Caption Construction}

We construct streaming caption supervision from event-rich videos through video curation, multi-granularity annotation, verification, and temporally aligned sequence conversion. The goal is to produce factual records that describe only evidence already available in the stream, while preserving complementary local and event-level information for later use.

\textbf{\textit{High-quality video curation.}}
We first build a diverse candidate pool from multiple video sources. Semantic retrieval is used to improve content coverage, while scene detection and visual-richness assessment identify videos with sufficient temporal variation and observable event content. We prioritize videos with clear visual changes, coherent temporal progression, and frequent actions or state transitions, since these properties provide useful supervision for incremental evidence recording. Videos with poor visual quality, prolonged static intervals, black frames, excessive shot fragmentation, or decoding failures are removed before annotation.

\textbf{\textit{Multi-granularity caption annotation.}}
For each selected video, we use Gemini3.1Pro and Seed2.1Pro to generate timestamp-grounded annotations at three temporal granularities. Frame- or clip-level captions describe local visual evidence, including visible objects, actions, scene details, state changes, and other temporally localized facts. Segment-level captions summarize coherent events spanning multiple local observations and capture their higher-level semantic progression, while video-level captions provide a global description of the full recording and model relations across segments. This hierarchy separates fine-grained evidence preservation from coarser event abstraction, allowing the same video to provide supervision at complementary temporal scales. We further verify the generated annotations against the source video before sequence construction. Local captions are checked for factual support and temporal alignment with their annotated intervals, while segment- and video-level captions are additionally checked for consistency with the events they summarize. Annotations containing unsupported content, incorrect temporal boundaries, or cross-level inconsistencies are discarded, preventing targets from introducing information before the required visual evidence becomes available.

\textbf{\textit{Causal streaming sequence construction.}}
The verified annotations are then converted into temporally ordered visual-language sequences. A frame- or clip-level caption is released as a \texttt{</Observe>} target only after the end of its supporting interval, so the corresponding local fact is generated after the required visual evidence has been observed. A segment-level caption is released with \texttt{</Summary>} when the associated event or segment is complete. All records are inserted according to their release times and become available only to subsequent predictions. The resulting sequence therefore interleaves incoming visual content with dense local-detail records and sparser event-level summaries while preserving the temporal order of evidence acquisition. The video-level caption is used differently from the two PHCM record types. After the video stream ends, we append a full-video summarization instruction and use the video-level annotation as the target following \texttt{</Response>}. It therefore provides instruction-conditioned full-video supervision rather than an additional persistent memory record. Together, these targets teach the model to describe observed evidence at multiple temporal scales and to release each output only when its supporting content has become available.

\begin{figure}[p]
    \centering
    \includegraphics[width=\textwidth,height=0.90\textheight,keepaspectratio]{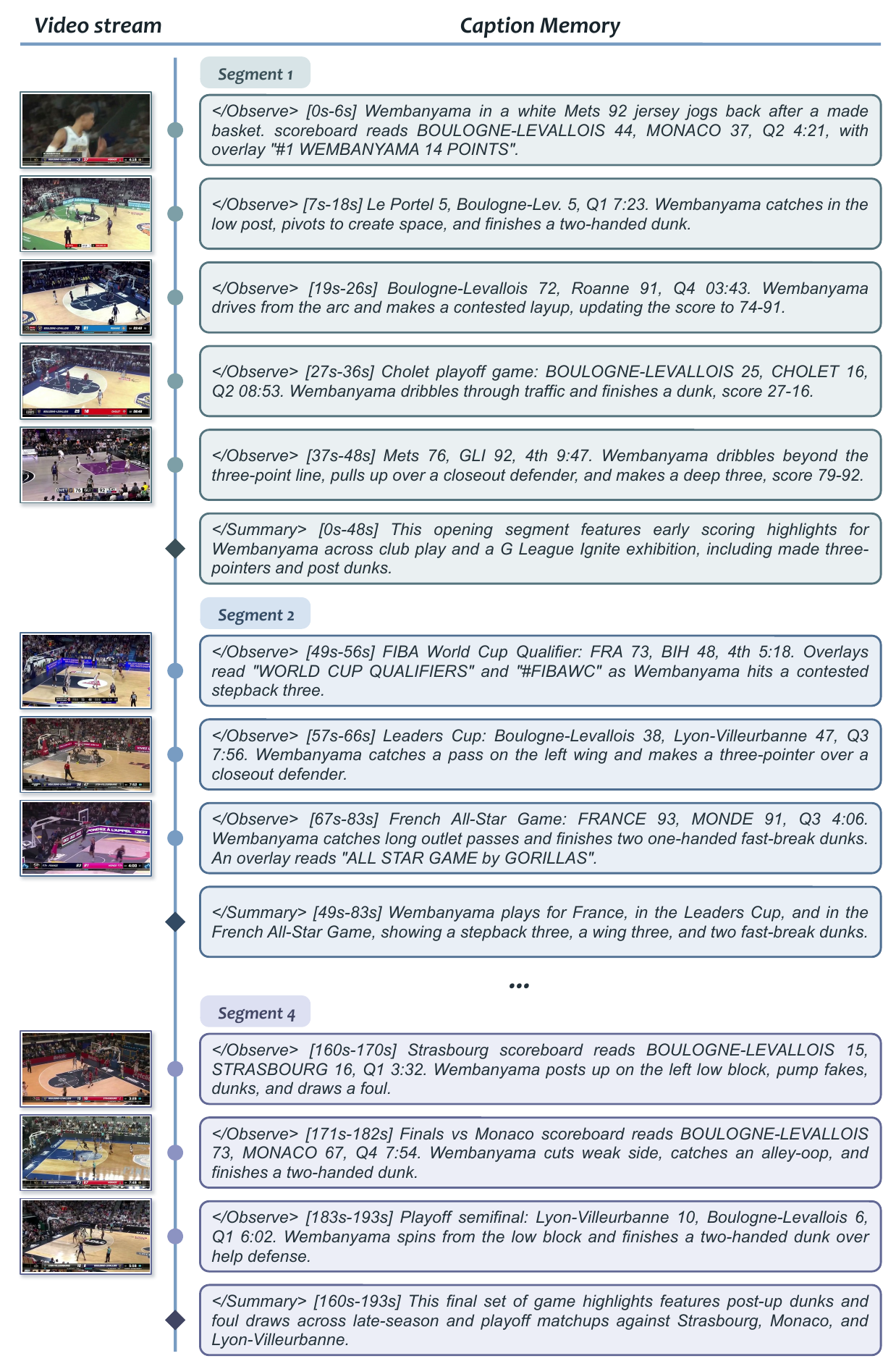}
    \caption{\textbf{Caption Memory training data from a sports video.} Closed-source models annotate local actions and scoreboard details as time-stamped \texttt{</Observe>} targets and completed segments as \texttt{</Summary>} targets. Selected segments are shown; the ellipsis indicates omitted content.}
    \label{fig:app:caption_memory_demo}
\end{figure}

\begin{figure}[p]
    \centering
    \includegraphics[width=\textwidth,height=0.90\textheight,keepaspectratio]{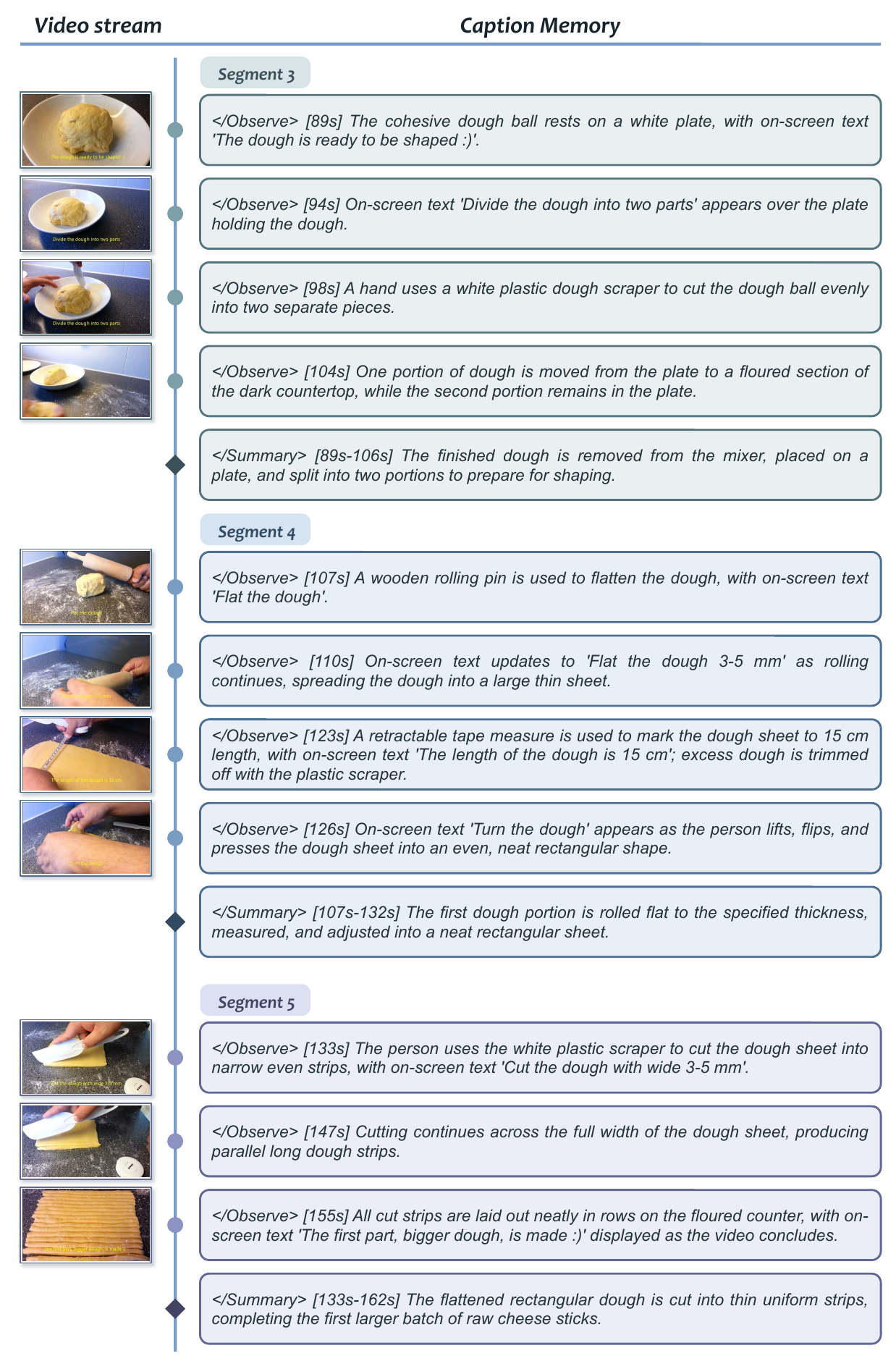}
    \caption{\textbf{Caption Memory training data from a cooking video.} Closed-source models provide time-stamped \texttt{</Observe>} targets for local actions and on-screen instructions, and \texttt{</Summary>} targets for completed segments. The selected segments show dough division, rolling and shaping, and cutting into strips.}
    \label{fig:app:caption_memory_demo_02}
\end{figure}

\begin{figure}[p]
    \centering
    \includegraphics[width=\textwidth,height=0.90\textheight,keepaspectratio]{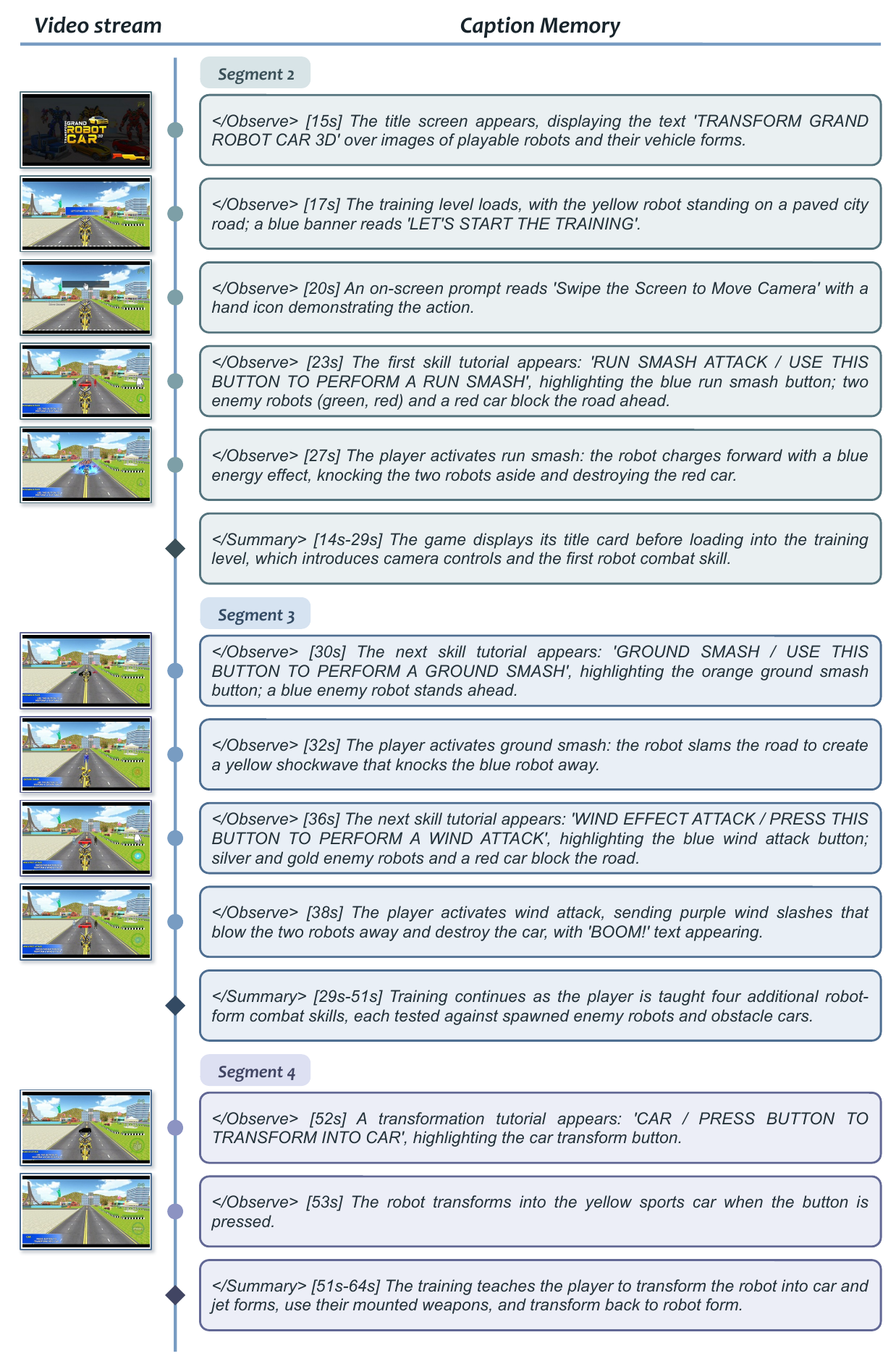}
    \caption{\textbf{Caption Memory training data from a gameplay video.} Closed-source models provide time-stamped \texttt{</Observe>} targets for on-screen instructions and player actions, and \texttt{</Summary>} targets for completed tutorial segments. Selected segments cover camera controls, combat skills, and robot-to-vehicle transformation.}
    \label{fig:app:caption_memory_demo_03}
\end{figure}

Figures~\ref{fig:app:caption_memory_demo},~\ref{fig:app:caption_memory_demo_02},~\ref{fig:app:caption_memory_demo_03} provide representative examples of the resulting Caption Memory training data across sports, cooking, and gameplay videos. Across these diverse domains, \texttt{</Observe>} targets preserve temporally localized facts such as actions, visible text, and state changes, while \texttt{</Summary>} targets consolidate completed events at a coarser temporal scale. These examples illustrate how the construction pipeline produces complementary local and event-level supervision for PHCM.

\clearpage

\clearpage

\clearpage

\clearpage

\subsection{Streaming QA Construction}

We construct streaming QA supervision through task-directed question-answer generation, coarse evidence localization, and fine-grained response-time calibration. The goal is to ensure that each response is grounded in identifiable visual evidence and assigned to an earlier point at which the available evidence becomes sufficient to support the answer.

\textbf{\textit{Task-directed streaming QA design.}}
We first collect structured metadata from two complementary sources: the verified streaming caption annotations described above and selected annotations from existing video datasets. We organize QA generation around target capabilities for streaming interaction, including state tracking, action prediction, proactive reminders, and proactive counting. For each capability, we combine the source metadata with a task-specific template that specifies the desired interaction pattern and guides the generation of questions and reference answers. Gemini3.1Pro and Seed2.1Pro then generate question--answer pairs grounded in the provided metadata and tailored to the corresponding streaming interaction setting. This capability-directed design yields QA pairs covering diverse forms of streaming understanding and response.

\textbf{\textit{Coarse evidence interval localization.}}
For each synthesized question--answer pair, we provide the source video, question, and reference answer to Seed2.1Pro and Gemini3.1Pro to localize a coarse temporal interval containing the visual evidence needed to support the answer. Because temporal relevance alone does not guarantee that an interval contains sufficient evidence to answer the question, we apply a separate verification step. We provide only the localized video clip and the user question to Qwen3.5-397B-A17B, without the reference answer or any visual context outside the predicted interval, and ask the model to answer the question. We retain the example only when the verifier can correctly recover the reference answer from this restricted evidence; otherwise, the QA pair is discarded. The verified interval therefore identifies a coarse region that contains sufficient supporting evidence, but it does not directly determine the response time.

\textbf{\textit{Fine-grained response-time calibration.}}
Within each verified coarse evidence interval, we evaluate valid integer-second candidate times at 1-s increments using preceding visual windows sampled at 4 fps. These windows contain only frames strictly before the candidate time and may extend before the interval start. For reasoning-oriented QA, Qwen3.5-9B scores the fixed reference answer conditioned on the question and a 4-s window. The score is the geometric mean of conditional token probabilities (inverse perplexity), obtained by exponentiating the mean token log probability. For perception-oriented tasks, WeMM-Embedding-9B~\cite{zhou2026wemm} separately encodes the target event description and the ordered frames of a 2-s window, then computes cosine similarity between their L2-normalized embeddings. Manual checks of randomly sampled reminder and QA examples supported these windows, as locally observable response cues can be brief even within long annotated intervals. For example, proactive reminder tasks often require the model to respond once a target appears; although the target may remain visible for tens of seconds, one or two seconds of observation are typically sufficient to trigger a timely response. To reduce sensitivity to isolated score fluctuations, both branches smooth each score using the median of the current and up to two preceding scores within the same candidate sequence. We compute the peak smoothed score over eligible candidates and select the earliest eligible candidate whose smoothed score is no more than 0.04 below this peak. Both branches use this additive margin as a fixed score tolerance, favoring earlier near-peak candidates.  A sole valid candidate is selected directly, while zero-length intervals retain their annotated timestamp without score-based recalibration. After determining the response time, we further validate its answerability by providing Qwen3.5-9B with the video prefix up to the selected candidate together with the user question, and discard all samples for which the model fails to produce the correct answer. This calibration advances the response time by 13.72 s on average relative to the original response-time annotation or the end of the evidence interval, with 86.62\% of records receiving an earlier response time. Manual inspection of randomly sampled examples suggests that, although our calibration does not always identify the earliest answerable moment, the selected response times generally contain sufficient visual evidence to support the answer, providing a favorable trade-off between response timeliness and evidence sufficiency.

Together, these stages convert task-directed question-answer pairs into evidence-grounded streaming training sequences with calibrated response times. By separating evidence localization from response-time calibration, the construction distinguishes where supporting evidence occurs from when it becomes sufficient for a response, providing temporally aligned supervision for proactive response.

\clearpage
\section{Implementation Details}
\label{sec:app:implementation}

\subsection{Training}

Table~\ref{tab:training_configuration} summarizes the SFT configuration. Initialization uses the instruction-tuned checkpoint with registered response-control and summary tokens. Loaded examples are counted after filtering and source sampling, before packing; the global batch size counts packed sequences per optimizer step. Both length limits include text and visual tokens. Frame sampling is dataset-specific, and annotation-specified dimensions override dynamic resizing. Cosine decay follows linear warmup. Training uses PyTorch FSDP, FlashAttention-2, and full activation recomputation on 32 NVIDIA H200 GPUs across four nodes. Epochs and optimizer steps denote the configured schedule.

\begin{table}[ht]
    \centering
    \setlength{\tabcolsep}{7pt}
    \renewcommand{\arraystretch}{1.1}
    \caption{\textbf{Parameter settings for supervised fine-tuning.}}
    \label{tab:training_configuration}

    \resizebox{0.52\textwidth}{!}{%
    \begin{tabular}{cl|c}
        \toprule
        & \textbf{Parameter} & \textbf{SFT} \\
        \midrule
        \multirow{4}{*}{\rotatebox[origin=c]{90}{\textbf{\textit{Data}}}}
        & Dataset & Image\&(Offline/Online)-Video \\
        & \# Loaded examples & 991,773 \\
        & \# Packed sequences & 150,261 \\
        & Sequence packing & Soft \\
        \midrule
        \multirow{3}{*}{\rotatebox[origin=c]{90}{\textbf{\textit{Model}}}}
        & Initialization & Qwen3-VL-4B \\
        & Trainable modules & projector\&LLM \\
        & Frozen module & Vision \\
        \midrule
        \multirow{3}{*}{\rotatebox[origin=c]{90}{\textbf{\textit{Vision}}}}
        & Sampling rate & Dynamic (1--4 FPS) \\
        & Maximum frames & 2,048 \\
        & Resolution & Dynamic \\
        \midrule
        \multirow{17}{*}{\rotatebox[origin=c]{90}{\textbf{\textit{Training}}}}
        & Optimizer & AdamW \\
        & Peak / minimum LR & $1\times10^{-5}$ / $1\times10^{-6}$ \\
        & LR schedule & Cosine \\
        & Warmup ratio & 0.03 \\
        & Adam $(\beta_1,\beta_2)$ & $(0.9,\,0.95)$ \\
        & Adam $\epsilon$ & $1\times10^{-8}$ \\
        & Weight decay & 0 \\
        & Max. gradient norm & 1.0 \\
        & Global batch size & 128 \\
        & Sample / pack length (tokens) & 131,072 / 131,072 \\
        & Epochs / planned steps & 1 / 1,174 \\
        & Precision & BF16 \\
        & Distributed training & FSDP \\
        & Recompute ratio & 1.0 \\
        & Attention backend & FA2 \\
        & GPUs & 32 \\
        & Random seed & 42 \\
        \bottomrule
    \end{tabular}%
    }
\end{table}

\subsection{Evaluation}

We run inference with Hugging Face Transformers and FlashAttention-2 through VLMEvalKit and dedicated task adapters. Model precision is set to BF16 or inferred from the checkpoint by the corresponding wrapper. Videos are sampled at 1--4 FPS and resized dynamically, preserving aspect ratio and satisfying the processor's spatial alignment requirements under the configured frame and pixel budgets. At each query or decision time, inputs contain only the available video clip or stream prefix, together with the task prompt and permitted dialogue history. Greedy decoding is the default and is implemented by explicitly setting \texttt{do\_sample=False}; sampling configurations retain their specified distribution parameters. The \texttt{max\_new\_tokens} setting limits each generation call independently. Each checkpoint is evaluated on eight GPUs by distributing independent samples across workers, with one or three worker processes per GPU according to the inference configuration. Outputs from these workers are merged before scoring. Prediction generation and metric computation are separate stages: configured local scorers operate on parsed answers or response timestamps.

\clearpage

\section{Per-benchmark Results}
\label{sec:app:benchmarks}

\textbf{Evaluation of comparison models.}
For comparison models with detailed results already reported on a benchmark, we directly use the metrics reported in the corresponding paper. For models without reported results on a given benchmark, we evaluate their officially released checkpoints; these re-evaluated models are marked with a dagger ($^\dagger$) in the tables. To avoid disadvantaging a comparison model due to differences in evaluation configuration, we evaluate each such model under two settings: the evaluation setting reported in its original paper and the same setting used for OneStreamer. For each model--benchmark pair, we report the complete set of results from the setting that achieves the higher benchmark-level average or aggregate score. The selection is made between the two evaluation runs at the whole-benchmark level, rather than independently for individual subtasks or metrics.

\textbf{Evaluation settings for OneStreamer.}
We use benchmark-specific inference configurations that follow the input and evaluation protocols of each benchmark. For OVOBench and StreamingBench, we construct hierarchical caption--summary memory using a 128-frame visual context sampled at 4\,fps, and answer each question using the accumulated memory together with the 16 most recent original-resolution frames sampled at 1\,fps. For OVBench and ODVBench, we sample the observed video segments at 2\,fps and 4\,fps, respectively. For the proactive-response benchmarks, we evaluate ProactiveVideoQA at 2\,fps and use Qwen3-235B-A22B as the automated judge. On OmniMMI, SG, AP, MD, and SI use the 32 most recent frames sampled at 4\,fps together with ASR transcripts from the corresponding observation window, whereas PA is evaluated without ASR at 1\,fps using a 32-frame window. OVO-Timing uses a 64-frame window sampled at 4\,fps, while ViSpeak-Bench uses a 32-frame window sampled at 1\,fps.

\begin{table}[ht]
    \centering
    \setlength{\tabcolsep}{2pt}
    \renewcommand{\arraystretch}{1.1}
    \caption{\textbf{Detailed evaluation results on OVO-Bench.} Each group average is the unweighted mean of its subtasks. Overall average is the unweighted mean of the three group averages.}
    \label{tab:app:ovobench}

    \resizebox{\textwidth}{!}{%
    \begin{tabular}{lc|cccccc|c|ccc|c|ccc|c|c}
        \toprule
        \multirow{2}{*}{\textbf{Method}} & \multirow{2}{*}{\textbf{Size}}
        & \multicolumn{7}{c|}{\textbf{Real-Time Visual Perception}}
        & \multicolumn{4}{c|}{\textbf{Backward Tracing}}
        & \multicolumn{4}{c|}{\textbf{Forward Active}}
        & \multirow{2}{*}{\textbf{Overall Avg.}} \\
        \cmidrule{3-17}
        & & OCR & ACR & ATR & STU & FPD & OJR & Avg.
        & EPM & ASI & HLD & Avg. & REC & SSR & CRR & Avg. & \\
        \midrule
        MOSS-VL-Realtime~\cite{wang2026mossvl} & 11B & - & - & - & - & - & - & 75.9 & - & - & - & 72.6 & - & - & - & \textbf{62.1} & 70.2 \\
        AURA~\cite{lu2026aura} & 8B & 89.9 & 79.8 & 80.2 & 70.2 & 77.2 & 81.5 & 79.8 & 54.9 & 67.6 & 58.6 & 60.4 & 30.4 & 75.8 & \textbf{61.3} & 55.8 & 65.3 \\
        JoyAI-VL-Interaction~\cite{yao2026joyaivlinteraction} & 8B & 80.5 & 60.6 & 77.6 & 65.2 & \textbf{79.2} & 62.0 & 70.9 & 58.9 & 62.8 & 23.1 & 48.3 & \textbf{45.8} & 70.3 & 58.8 & 58.3 & 59.1 \\
        Mage-VL$^\dagger$~\cite{yang2026magevl} & 4B & \textbf{96.0} & \textbf{81.7} & \textbf{86.2} & \textbf{71.3} & 72.3 & \textbf{82.6} & \textbf{81.7} & 53.2 & 54.7 & 38.7 & 48.9 & 19.5 & 73.3 & 41.7 & 44.8 & 58.5 \\
        VideoChat3~\cite{li2026videochat3} & 4B & 85.9 & 71.6 & 79.3 & 55.1 & 73.3 & 71.2 & 72.7 & 56.9 & 58.1 & 47.3 & 54.1 & 23.8 & 67.2 & 55.0 & 48.7 & 58.5 \\
        Qwen3-VL$^\dagger$~\cite{bai2025qwen3vl} & 4B & 91.3 & 67.9 & 77.6 & 59.6 & 75.2 & 67.4 & 73.2 & 59.9 & 59.5 & 41.9 & 53.8 & 34.2 & 61.5 & 52.5 & 49.4 & 58.8 \\
        \rowcolor[HTML]{DCEEFC}
        \textbf{\ours} & 4B & 94.6 & 80.7 & 81.9 & 70.2 & 78.2 & \textbf{82.6} & 81.4 & \textbf{62.6} & \textbf{71.6} & \textbf{99.5} & \textbf{77.9} & 35.5 & \textbf{78.5} & 56.7 & 56.9 & \textbf{72.1} \\
        \bottomrule
    \end{tabular}%
    }
\end{table}

\FloatBarrier

\begin{table}[ht]
\setlength{\tabcolsep}{2.5pt}

\renewcommand{\arraystretch}{1.1}
\centering
\caption{\textbf{Detailed results on the real-time understanding tasks of StreamingBench.} ALL is accuracy over all 2,498 questions.}

\resizebox{\textwidth}{!}{%
\begin{tabular}{lc|cccccccccc|c}
\toprule
\textbf{Method} & \textbf{Size} & OP & CR & CS & ATP & EU & TR & PR & SU & ACP & CT & \textbf{ALL} \\ \midrule
        MOSS-VL-Realtime~\cite{wang2026mossvl} & 11B & - & - & - & - & - & - & - & - & - & - & 82.9 \\
        AURA~\cite{lu2026aura} & 8B & 87.5 & \textbf{84.4} & 93.4 & 89.5 & 83.2 & 87.9 & 86.1 & 76.4 & 82.4 & 47.7 & 83.2 \\
        JoyAI-VL-Interaction$^\dagger$~\cite{yao2026joyaivlinteraction} & 8B & 87.2 & 83.6 & 92.1 & 88.9 & 81.9 & 91.3 & \textbf{88.0} & 74.0 & 81.3 & 46.1 & 82.7 \\
        Mage-VL$^\dagger$~\cite{yang2026magevl} & 4B & 88.3 & 71.9 & 91.2 & 89.8 & 75.6 & 90.7 & 83.3 & \textbf{81.7} & 81.9 & 40.9 & 82.1 \\
        VideoChat3~\cite{li2026videochat3} & 4B & 86.4 & \textbf{84.4} & 93.7 & 88.5 & 81.3 & 83.8 & 78.7 & 78.9 & 83.6 & 42.0 & 81.9 \\
        Qwen3-VL$^\dagger$~\cite{bai2025qwen3vl} & 4B & 83.7 & 78.9 & 91.8 & 87.9 & 83.1 & 93.8 & \textbf{88.0} & 75.6 & 76.2 & 47.7 & 81.8 \\
        \rowcolor[HTML]{DCEEFC}
        \textbf{\ours} & 4B & \textbf{90.7} & 80.5 & \textbf{95.9} & \textbf{93.1} & \textbf{84.4} & \textbf{94.1} & 87.0 & 80.1 & \textbf{85.3} & \textbf{61.7} & \textbf{86.9} \\
        \bottomrule
\end{tabular}%
}

\label{tab:app:streamingbench}

\end{table}

\FloatBarrier

\begin{table}[ht]
\setlength{\tabcolsep}{2pt}
    \centering
    \renewcommand{\arraystretch}{1.1}
\caption{\textbf{Detailed evaluation results on OVBench.} AVG is the unweighted mean of the 16 subtasks.}

    \resizebox{\textwidth}{!}{%
    \begin{tabular}{lc|cccccccccccccccc|c}
        \toprule
        \multirow{2}{*}{\textbf{Method}} & \multirow{2}{*}{\textbf{Size}} & \multicolumn{3}{c}{\textbf{FP}} & \multicolumn{3}{c}{\textbf{THV}} & \multicolumn{3}{c}{\textbf{PM}} & \multicolumn{2}{c}{\textbf{SP}} & \multicolumn{2}{c}{\textbf{STP}} & \multicolumn{3}{c|}{\textbf{TP}} & \multirow{2}{*}{\textbf{AVG}} \\
        \cmidrule(lr){3-5} \cmidrule(lr){6-8} \cmidrule(lr){9-11} \cmidrule(lr){12-13} \cmidrule(lr){14-15} \cmidrule(lr){16-18}
        & & AA & GSP & MP & AP & SV & OP & AR & PR & TR & AL & OP & AT & OT & AS & SL & OES & \\
        \midrule
        MOSS-VL-Realtime$^\dagger$~\cite{wang2026mossvl} & 11B & 52.1 & 56.1 & \textbf{63.5} & 43.0 & \textbf{70.4} & 40.0 & \textbf{70.5} & 66.1 & 12.9 & \textbf{55.4} & 55.0 & 60.8 & 34.7 & 69.0 & \textbf{75.3} & 34.1 & 53.7 \\
        AURA$^\dagger$~\cite{lu2026aura} & 8B & \textbf{67.9} & \textbf{76.9} & 26.2 & 50.2 & 58.0 & 72.7 & 30.3 & 71.8 & 65.9 & 39.7 & 57.2 & 93.9 & \textbf{58.8} & 76.3 & 26.4 & 61.2 & 58.3 \\
        JoyAI-VL-Interaction~\cite{yao2026joyaivlinteraction} & 8B & 55.1 & 54.0 & 29.8 & 48.1 & 51.3 & \textbf{76.6} & 50.0 & \textbf{73.6} & 87.1 & 35.0 & \textbf{89.5} & 94.8 & 56.5 & 73.0 & 32.3 & 90.2 & 62.3 \\
        Mage-VL$^\dagger$~\cite{yang2026magevl} & 4B & 53.8 & 67.0 & 34.8 & \textbf{68.2} & 57.7 & 73.0 & 37.8 & 61.5 & 76.7 & 43.8 & 61.6 & 70.4 & 56.0 & 73.5 & 27.4 & 56.5 & 57.5 \\
        VideoChat3~\cite{li2026videochat3} & 4B & 59.0 & 62.0 & 23.3 & 63.9 & 58.2 & 69.4 & 46.3 & 68.7 & 79.6 & 35.6 & 85.9 & 95.2 & 58.1 & 81.9 & 26.5 & 85.6 & 62.5 \\
        Qwen3-VL$^\dagger$~\cite{bai2025qwen3vl} & 4B & 55.1 & 62.7 & 25.8 & 59.2 & 56.8 & 69.4 & 34.7 & 71.4 & 69.9 & 34.4 & 56.2 & 95.7 & 53.1 & 74.9 & 27.3 & 39.6 & 55.4 \\
        \rowcolor[HTML]{DCEEFC}
        \textbf{\ours} & 4B & 57.7 & 74.4 & 24.0 & 60.1 & 55.7 & 68.3 & 67.8 & 69.5 & \textbf{88.9} & 48.4 & \textbf{89.5} & \textbf{99.1} & 58.1 & \textbf{86.0} & 29.8 & \textbf{91.0} & \textbf{66.8} \\
        \bottomrule
\end{tabular}%
    }

\label{tab:app:ovbench}
    \end{table}

\FloatBarrier

\begin{table}[!ht]
    \centering
    \setlength{\tabcolsep}{2pt}
    \renewcommand{\arraystretch}{1.1}
    \caption{\textbf{Detailed evaluation results on ODV-Bench.} Group Avg and Overall scores retain the original sample weighting.}
    \label{tab:app:odvbench}

    \resizebox{\textwidth}{!}{%
    \begin{tabular}{lc|cccccc|c|ccc|c|ccc|c|c}
        \toprule
        \multirow{2}{*}{\textbf{Method}} &
        \multirow{2}{*}{\textbf{Size}} &
        \multicolumn{7}{c|}{\textbf{Static Target}} &
        \multicolumn{4}{c|}{\textbf{Dynamic Target}} &
        \multicolumn{4}{c|}{\textbf{Event Oriented}} &
        \multirow{2}{*}{\textbf{Overall}} \\
        \cmidrule{3-17}
        & & RTP & HD & KIE & TCD & DDM & PTM & Avg.
        & AP & LP & DP & Avg. & RP & RA & ARA & Avg. & \\
        \midrule
        MOSS-VL-Realtime$^\dagger$~\cite{wang2026mossvl} & 11B & \textbf{75.3} & \textbf{99.2} & 84.9 & 49.1 & 39.6 & 79.1 & \textbf{71.3} & 65.7 & 85.6 & 48.5 & 62.6 & 50.5 & 73.4 & 53.9 & 59.2 & 63.9 \\
        AURA$^\dagger$~\cite{lu2026aura} & 8B & 59.1 & 77.2 & 50.9 & \textbf{74.5} & 39.9 & 58.7 & 57.0 & 45.8 & 74.7 & 50.9 & 56.2 & 55.0 & 80.2 & \textbf{63.5} & 65.0 & 58.8 \\
        JoyAI-VL-Interaction~\cite{yao2026joyaivlinteraction} & 8B & 70.8 & 52.0 & \textbf{90.6} & 47.3 & 40.3 & 78.7 & 66.4 & \textbf{80.1} & 96.1 & 60.4 & 74.7 & 39.4 & 91.7 & 57.1 & 60.1 & 68.6 \\
        Mage-VL$^\dagger$~\cite{yang2026magevl} & 4B & 68.0 & 37.4 & 64.2 & \textbf{74.5} & \textbf{46.9} & 77.4 & 65.2 & 58.9 & 76.1 & 52.6 & 60.5 & \textbf{68.2} & 75.4 & 51.3 & 69.3 & 64.1 \\
        VideoChat3~\cite{li2026videochat3} & 4B & 70.4 & 87.8 & 73.6 & \textbf{74.5} & 32.0 & \textbf{85.5} & 70.2 & 74.4 & 93.4 & 56.5 & 70.7 & 60.4 & 90.7 & 62.8 & \textbf{71.7} & 70.8 \\
        Qwen3-VL$^\dagger$~\cite{bai2025qwen3vl} & 4B & 57.6 & 31.7 & 67.9 & 47.3 & 31.4 & 67.8 & 54.4 & 63.7 & 78.6 & 49.1 & 60.5 & 42.1 & 74.2 & 61.5 & 55.6 & 57.6 \\
        \rowcolor[HTML]{DCEEFC}
        \textbf{\ours} & 4B & 68.5 & 90.2 & 84.9 & 60.0 & 28.4 & 84.0 & 68.4 & 76.8 & \textbf{96.5} & \textbf{64.4} & \textbf{76.0} & 50.8 & \textbf{92.0} & 50.0 & 65.8 & \textbf{71.3} \\
        \bottomrule
    \end{tabular}%
    }
\end{table}

\FloatBarrier

\begin{table}[!ht]
    \centering
    \setlength{\tabcolsep}{3pt}
    \renewcommand{\arraystretch}{1.1}
    \caption{\textbf{Detailed evaluation results on ProactiveVideoQA.} PAUC scores at $\omega=0.0$, $0.5$, and $1.0$. Avg. is the unweighted four-task mean at $\omega=0.5$.}
    \label{tab:app:proactivevideoqa}

    \resizebox{\textwidth}{!}{%
    \begin{tabular}{lc|ccc|ccc|ccc|ccc|c}
        \toprule
        \multirow{2}{*}{\textbf{Method}} & \multirow{2}{*}{\textbf{Size}} & \multicolumn{3}{c|}{\textbf{[WEB]}}
        & \multicolumn{3}{c|}{\textbf{[EGO]}}
        & \multicolumn{3}{c|}{\textbf{[TV]}}
        & \multicolumn{3}{c|}{\textbf{[VAD]}}
        & \multirow{2}{*}{\textbf{Avg.}} \\
        \cmidrule{3-14}
        & & 0.0 & 0.5 & 1.0 & 0.0 & 0.5 & 1.0 & 0.0 & 0.5 & 1.0 & 0.0 & 0.5 & 1.0 & \\
        \midrule
        MOSS-VL-Realtime~\cite{wang2026mossvl} & 11B & - & \textbf{55.4} & - & - & 47.8 & - & - & \textbf{50.2} & - & - & 35.3 & - & 47.2 \\
        AURA$^\dagger$~\cite{lu2026aura} & 8B & 34.7 & 40.3 & 45.8 & 25.8 & 26.0 & 26.3 & 29.1 & 31.3 & 33.5 & 23.4 & 25.6 & 27.8 & 30.8 \\
        JoyAI-VL-Interaction$^\dagger$~\cite{yao2026joyaivlinteraction} & 8B & 33.0 & 36.9 & 40.7 & 26.7 & 27.2 & 27.7 & 29.5 & 30.6 & 31.7 & 24.2 & 23.8 & 23.4 & 29.6 \\
        Mage-VL$^\dagger$~\cite{yang2026magevl} & 4B & 26.6 & 27.1 & 27.5 & 24.9 & 24.9 & 24.9 & 25.0 & 25.1 & 25.1 & 25.0 & 25.4 & 25.7 & 25.6 \\
        VideoChat3~\cite{li2026videochat3} & 4B & 36.0 & 39.6 & 43.3 & 30.3 & 32.2 & 34.2 & 35.2 & 39.6 & 44.0 & 34.1 & 38.9 & 43.7 & 37.6 \\
        Qwen3-VL$^\dagger$~\cite{bai2025qwen3vl} & 4B & 38.5 & 42.4 & 46.3 & 25.0 & 25.0 & 25.0 & 37.8 & 40.4 & 43.1 & 28.7 & 29.6 & 30.4 & 34.3 \\
        \rowcolor[HTML]{DCEEFC}
        \textbf{\ours} & 4B & \textbf{47.0} & 54.1 & \textbf{61.2} & \textbf{49.2} & \textbf{53.4} & \textbf{57.6} & \textbf{40.3} & 44.6 & \textbf{48.8} & \textbf{38.4} & \textbf{42.8} & \textbf{47.2} & \textbf{48.7} \\
        \bottomrule
    \end{tabular}%
    }
\end{table}

\FloatBarrier

\begin{table}[!ht]
    \centering
    \setlength{\tabcolsep}{8pt}
    \renewcommand{\arraystretch}{1.1}
    \caption{\textbf{Detailed evaluation results on OmniMMI.} All is the unweighted mean of the five tasks.}
    \label{tab:app:omnimmi}

    \resizebox{1\textwidth}{!}{%
    \begin{tabular}{lc|ccccc|c}
        \toprule
        \textbf{Method} & \textbf{Size} & \textbf{SG} & \textbf{AP} & \textbf{MD} & \textbf{SI} & \textbf{PA} & \textbf{All} \\
        \midrule
        MOSS-VL-Realtime~\cite{wang2026mossvl} & 11B & 21.7 & 33.5 & 10.7 & 31.5 & \textbf{66.0} & 32.7 \\
        AURA~\cite{lu2026aura} & 8B & \textbf{24.0} & 32.0 & 7.7 & 26.0 & 37.5 & 25.4 \\
        JoyAI-VL-Interaction$^\dagger$~\cite{yao2026joyaivlinteraction} & 8B & 9.7 & 25.0 & 5.0 & 33.5 & 16.0 & 17.8 \\
        Mage-VL$^\dagger$~\cite{yang2026magevl} & 4B & 9.3 & 30.5 & 2.0 & 30.5 & 5.5 & 15.6 \\
        VideoChat3$^\dagger$~\cite{li2026videochat3} & 4B & 13.0 & 31.5 & 7.0 & 48.5 & 23.0 & 24.6 \\
        Qwen3-VL$^\dagger$~\cite{bai2025qwen3vl} & 4B & 15.3 & 32.5 & \textbf{12.3} & 45.5 & 41.5 & 29.4 \\
        \rowcolor[HTML]{DCEEFC}
        \textbf{\ours} & 4B & 21.0 & \textbf{38.5} & \textbf{12.3} & \textbf{61.0} & 50.0 & \textbf{36.6} \\
        \bottomrule
    \end{tabular}%
    }
\end{table}

\FloatBarrier

\begin{table}[!ht]
    \centering
    \setlength{\tabcolsep}{8pt}
    \renewcommand{\arraystretch}{1.1}
    \caption{\textbf{Detailed evaluation results on OVO-Timing.} Proactive response timing on OVO-Bench Future Active Responding tasks. Avg. F1 is the unweighted mean over CRR, SSR, and REC.}
    \label{tab:app:ovotiming}

    \resizebox{\textwidth}{!}{%
    \begin{tabular}{lc|ccc|ccc|ccc|c}
        \toprule
        \multirow{2}{*}{\textbf{Method}} & \multirow{2}{*}{\textbf{Size}}
        & \multicolumn{3}{c|}{\textbf{CRR}}
        & \multicolumn{3}{c|}{\textbf{SSR}}
        & \multicolumn{3}{c|}{\textbf{REC}}
        & \multirow{2}{*}{\textbf{Avg. F1}} \\
        \cmidrule(lr){3-5} \cmidrule(lr){6-8} \cmidrule(lr){9-11}
        & & R & P & F1 & R & P & F1 & R & P & F1 & \\
        \midrule
        MOSS-VL-Realtime$^\dagger$~\cite{wang2026mossvl} & 11B & 39.6 & 21.2 & 27.6 & 46.4 & 21.6 & 29.4 & 54.2 & \textbf{63.2} & 58.4 & 38.5 \\
        AURA$^\dagger$~\cite{lu2026aura} & 8B & \textbf{70.8} & 3.0 & 5.8 & 13.5 & 5.5 & 7.8 & 39.4 & 13.2 & 19.8 & 11.1 \\
        JoyAI-VL-Interaction$^\dagger$~\cite{yao2026joyaivlinteraction} & 8B & 52.1 & 8.5 & 14.6 & 23.6 & 5.8 & 9.3 & 64.6 & 25.2 & 36.2 & 20.0 \\
        Mage-VL$^\dagger$~\cite{yang2026magevl} & 4B & 41.7 & 8.7 & 14.5 & \textbf{54.2} & 11.2 & 18.6 & 38.3 & 27.3 & 31.9 & 21.7 \\
        VideoChat3~\cite{li2026videochat3} & 4B & 35.4 & \textbf{28.1} & \textbf{31.4} & 43.7 & 16.1 & 23.5 & 45.6 & 46.1 & 45.8 & 33.6 \\
        Qwen3-VL$^\dagger$~\cite{bai2025qwen3vl} & 4B & 62.5 & 16.6 & 26.2 & 43.8 & 12.9 & 19.9 & 71.4 & 30.0 & 42.2 & 29.4 \\
        \rowcolor[HTML]{DCEEFC}
        \textbf{\ours} & 4B & 35.4 & 18.9 & 24.7 & 53.0 & \textbf{31.8} & \textbf{39.8} & \textbf{79.8} & 48.4 & \textbf{60.2} & \textbf{41.6} \\
        \bottomrule
    \end{tabular}%
    }
\end{table}

\FloatBarrier

\begin{table}[!ht]
    \centering
    \setlength{\tabcolsep}{2pt}
    \renewcommand{\arraystretch}{1.1}
    \caption{\textbf{Detailed evaluation results on ViSpeak-Bench.} Time All and Text All are unweighted means across the six tasks. Overall averages the per-task timing-gated text scores (zero for incorrect timing).}
    \label{tab:app:vispeak}

    \resizebox{\textwidth}{!}{%
    \begin{tabular}{lc|ccccccc|ccccccc|c}
        \toprule
        \multirow{2}{*}{\textbf{Method}} & \multirow{2}{*}{\textbf{Size}}
        & \multicolumn{7}{c|}{\textbf{Time Accuracy (\%)}}
        & \multicolumn{7}{c|}{\textbf{Text Score}}
        & \multirow{2}{*}{\textbf{Overall}} \\
        \cmidrule(lr){3-9} \cmidrule(lr){10-16}
        & & AW & VI & HR & VW & VT & GU & \textbf{All}
        & AW & VI & HR & VW & VT & GU & \textbf{All} & \\
        \midrule
        Moss-VL-Realtime$^\dagger$~\cite{wang2026mossvl} & 11B & 54.00 & 66.00 & 39.00 & 95.00 & 87.00 & \textbf{99.00} & 73.33 & 3.44 & 1.71 & 1.64 & 4.98 & \textbf{4.89} & 2.31 & 3.16 & 2.48 \\
        AURA$^\dagger$~\cite{lu2026aura} & 8B & 34.00 & 22.00 & 2.00 & 0.00 & 78.00 & 4.00 & 23.33 & 2.47 & 2.95 & 1.50 & 0.00 & 4.56 & 4.13 & 2.60 & 0.87 \\
        JoyAI-VL-Interaction$^\dagger$~\cite{yao2026joyaivlinteraction} & 8B & 19.00 & 16.00 & 60.00 & \textbf{97.00} & 57.00 & 92.50 & 56.92 & 3.05 & \textbf{4.63} & 1.67 & \textbf{5.00} & 4.56 & 3.48 & \textbf{3.73} & 2.16 \\
        Mage-VL$^\dagger$~\cite{yang2026magevl} & 4B & \textbf{55.00} & 65.00 & \textbf{82.00} & 55.00 & 26.00 & 56.00 & 56.50 & 1.60 & 0.08 & 0.41 & 3.91 & 0.00 & 3.00 & 1.50 & 0.85 \\
        VideoChat3$^\dagger$~\cite{li2026videochat3} & 4B & 13.50 & 1.00 & 1.00 & 58.00 & 26.00 & 71.50 & 28.50 & 1.52 & 0.00 & \textbf{3.00} & 4.95 & 4.62 & 2.79 & 2.81 & 1.05 \\
        Qwen3-VL$^\dagger$~\cite{bai2025qwen3vl} & 4B & 40.50 & 49.00 & 20.00 & 83.00 & 90.00 & 84.00 & 61.08 & \textbf{4.10} & 2.84 & 2.50 & \textbf{5.00} & 4.47 & 3.26 & 3.69 & 2.41 \\
        \rowcolor[HTML]{DCEEFC}
        \textbf{\ours} & 4B & 44.50 & \textbf{76.00} & 40.00 & 89.00 & \textbf{99.00} & 98.50 & \textbf{74.50} & 3.29 & 4.09 & 2.20 & 4.93 & 3.06 & \textbf{4.42} & 3.67 & \textbf{2.87} \\
        \bottomrule
    \end{tabular}%
    }
\end{table}

\FloatBarrier

\clearpage

\section{Extended Ablations}

\subsection{Effect of Proactive Caption Supervision}
\label{sec:app:caption_supervision}

\begin{table}[!ht]
    \centering
    \renewcommand{\arraystretch}{1.2}
    \setlength{\tabcolsep}{8pt}
    \caption{\textbf{Effect of proactive caption supervision under visual-only evaluation.}
    All settings use visual inputs without caption memory. Full uses the complete observed video history sampled at 4 fps, while FIFO retains the 16 most recent frames sampled at 1 fps.}
    \label{tab:caption_data_transfer}

    \resizebox{\textwidth}{!}{%
    \begin{tabular}{l|cc|cc|c|c}
        \toprule
        \multirow{2}{*}{\textbf{Model}} & \multicolumn{2}{c|}{\textbf{OVOBench}} & \multicolumn{2}{c|}{\textbf{StreamingBench}} & \multirow{2}{*}{\textbf{OVBench}} & \multirow{2}{*}{\textbf{ODVBench}} \\
        \cmidrule(lr){2-3}\cmidrule(lr){4-5}
        & \textbf{Full} & \textbf{FIFO} & \textbf{Full} & \textbf{FIFO} & & \\
        \midrule
        w/o proactive caption data & 69.8 & 70.0 & 80.1 & 85.8 & 65.7 & 70.5 \\
        \rowcolor[HTML]{DCEEFC}
        \textbf{\ours} & \textbf{70.9} & \textbf{70.1} & \textbf{80.5} & \textbf{86.3} & \textbf{66.8} & \textbf{71.3} \\
        \bottomrule
    \end{tabular}%
    }
\end{table}

To examine whether proactive caption supervision benefits video understanding beyond inference-time memory use, we compare {\ours} with a variant trained without the 60K proactive caption examples in Table~\ref{tab:caption_data_transfer}. Both models are evaluated using visual inputs without caption memory. {\ours} achieves higher aggregate scores across all six evaluation settings, covering full-history and recent-window QA on OVOBench and StreamingBench as well as OVBench and ODVBench. These gains support the role of streaming caption targets as fine-grained supervision for interpreting observed video prefixes. Thus, proactive caption supervision contributes to perception learning even when generated records are not used at inference.

\begin{table}[!ht]
    \centering
    \renewcommand{\arraystretch}{1.2}
    \setlength{\tabcolsep}{8pt}
    \caption{
        \textbf{Effect of proactive caption supervision on inference-time memory use.}
        FIFO uses only the recent visual window, while PHCM retains each model's own proactively generated descriptions for subsequent QA under the same fact-recording prompt and inference settings. $\Delta$ denotes PHCM minus FIFO in percentage points.
    }
    \label{tab:caption_data_memory}

    \resizebox{\textwidth}{!}{%
    \begin{tabular}{l|ccc|ccc}
        \toprule
        \multirow{2}{*}{\textbf{Model}} & \multicolumn{3}{c|}{\textbf{OVOBench}} & \multicolumn{3}{c}{\textbf{StreamingBench}} \\
        \cmidrule(lr){2-4}\cmidrule(lr){5-7}
        & \textbf{FIFO} & \textbf{PHCM} & $\boldsymbol{\Delta}$ & \textbf{FIFO} & \textbf{PHCM} & $\boldsymbol{\Delta}$ \\
        \midrule
        w/o proactive caption data & 70.0 & 69.9 & \textcolor{red!75!black}{$-0.1$} & 85.8 & 86.1 & \textcolor{green!50!black}{$+0.3$} \\
        \rowcolor[HTML]{DCEEFC}
        \textbf{\ours} & \textbf{70.1} & \textbf{72.1} & \textcolor{green!50!black}{$\boldsymbol{+2.0}$} & \textbf{86.3} & \textbf{86.9} & \textcolor{green!50!black}{$\boldsymbol{+0.6}$} \\
        \bottomrule
    \end{tabular}%
    }
\end{table}

Table~\ref{tab:caption_data_memory} further examines whether proactive caption supervision improves the effectiveness of memory construction and use. Both models use the same proactive fact-recording prompt and inference settings. Each model's generated descriptions are retained for subsequent QA. Since the ablated model is trained without the 60K proactive caption examples and therefore never sees \texttt{</Observe>} or \texttt{</Summary>} during training, it describes the observed content through \texttt{</Response>} instead. These response-form descriptions are likewise retained as memory. Adding PHCM to the same recent visual window improves {\ours} by 2.0 percentage points on OVOBench Overall and 0.6 points on StreamingBench Real-Time. In contrast, retaining the ablated model's descriptions yields changes of $-0.1$ and $+0.3$ points relative to FIFO, respectively. These results suggest that inference-time prompting alone does not reproduce the memory gains obtained with dedicated caption supervision. Such supervision helps the model learn to construct and use reusable factual memory from streaming observations.

\subsection{Effectiveness of Hierarchical Caption Memory}

\begin{table}[ht]
    \centering
    \renewcommand{\arraystretch}{1.2}
    \setlength{\tabcolsep}{6pt}

    \caption{
        \textbf{Ablation of hierarchical Caption Memory.} We evaluate the contributions of local-detail and event-level memory records when retained individually or together.
    }

    \resizebox{1\textwidth}{!}{
        \begin{tabular}{lcc|c|c}
            \toprule
            \textbf{Memory Variant} & \textbf{\texttt{</Observe>}} & \textbf{\texttt{</Summary>}} & \textbf{StreamingBench (Real-Time)} & \textbf{OVOBench (Overall)} \\
            \midrule

        No Caption Memory & $\times$ & $\times$ & 86.3 & 70.1 \\

        Observe only & $\checkmark$ & $\times$ & \textbf{86.9} & 71.7 \\

        Summary only & $\times$ & $\checkmark$ & 86.6 & 71.2 \\

        \rowcolor[HTML]{DCEEFC}
        \textbf{Full Hierarchy} & $\checkmark$ & $\checkmark$ & \textbf{86.9} & \textbf{72.1} \\

            \bottomrule
        \end{tabular}
    }

    \label{tab:ablation_caption_memory_levels}
\end{table}

Table~\ref{tab:ablation_caption_memory_levels} isolates the contributions of the two record types in PHCM by retaining local-detail \texttt{</Observe>} captions and event-level \texttt{</Summary>} records individually or together. Compared with using no Caption Memory, retaining only \texttt{</Observe>} records improves StreamingBench Real-Time from 86.3 to 86.9 and OVOBench Overall from 70.1 to 71.7. Retaining only \texttt{</Summary>} records also improves both metrics, reaching 86.6 on StreamingBench and 71.2 on OVOBench. Combining both record types yields the highest OVOBench Overall score of 72.1 while matching the best StreamingBench Real-Time score of 86.9. These results indicate that fine-grained observation records provide most of the individual gain, while event-level summaries contribute complementary information that further improves overall online understanding when the two are combined.

\subsection{Efficiency of Online Memory Construction}
\label{sec:app:memory_construction_efficiency}

\begin{table}[ht]
    \centering
    \renewcommand{\arraystretch}{1.2}
    \setlength{\tabcolsep}{4pt}
    \caption{\textbf{Efficiency of online memory construction.}
        We replay a 360\,s StreamingBench clip at one update per second on a single NVIDIA H200.
        FIFO and PHCM use a 32\,s visual window sampled at 4 FPS, while Full retains all observed frames at 4 FPS.
        FIFO and Full are forced-silence controls, whereas PHCM constructs caption memory online.
    }
    \label{tab:online_memory_construction_efficiency}
    \resizebox{\textwidth}{!}{%
        \begin{tabular}{lcccc}
            \toprule
            \textbf{Strategy} & \textbf{Processing time (s)} & \textbf{Mean update (s)} & \textbf{Peak memory (GiB)} & \textbf{Peak input tokens} \\
            \midrule
            FIFO (forced silence) & 170.00 & 0.472 & 10.233 & 7,914 \\
            Full (forced silence) & 1444.75 & 4.013 & 29.754 & 87,535 \\
            \rowcolor[HTML]{DCEEFC}
            \textbf{PHCM (Ours)} & 228.96 & 0.636 & 10.327 & 8,362 \\
            \bottomrule
        \end{tabular}%
    }
\end{table}

To evaluate whether PHCM can construct memory online without falling behind the incoming video stream, we conduct a streaming throughput test under a relatively demanding visual load. We replay a 360\,s StreamingBench clip at one update per second on a single NVIDIA H200. FIFO and PHCM use a 32\,s recent visual window sampled at 4 FPS, corresponding to 128 frames per update. Full instead retains all observed frames from the beginning of the stream, also sampled at 4 FPS. This setting tests both the additional cost of online memory writing and the scalability of PHCM relative to continuously retaining visual history.

For FIFO and Full, we use forced-silence controls that follow the same update-by-update inference pipeline but emit only \texttt{</Silence>} and the end-of-turn token. PHCM follows the same update schedule while generating \texttt{</Observe>} and \texttt{</Summary>} records when triggered. All three settings invoke the model at every update and retain the resulting textual history.

As shown in Table~\ref{tab:online_memory_construction_efficiency}, PHCM averages 0.636\,s per update, remaining below the 1\,s input interval even with a 128-frame visual context. The measurement includes video decoding, preprocessing, input construction, vision encoding, and language-model inference. Over the 360\,s clip, PHCM generates 26 local-detail observations and two event summaries. Compared with the FIFO forced-silence control, online memory construction increases the mean update latency from 0.472\,s to 0.636\,s, while peak GPU memory increases only from 10.233 to 10.327\,GiB and peak input length from 7,914 to 8,362 tokens.

In contrast, continuously retaining the full visual history increases the mean update latency to 4.013\,s, well beyond the 1\,s update interval, with 29.754\,GiB peak GPU memory and 87,535 peak input tokens. PHCM therefore reduces mean update latency by 84.2\%, peak GPU memory by 65.3\%, and peak input length by 90.4\% relative to Full. Together, these results show that PHCM can construct reusable memory online with modest additional cost over a recent-window baseline, while avoiding the rapidly growing context and computation required to retain the complete visual history.

\subsection{Effectiveness of Training Data Construction}
\label{sec:app:training_data}

\begin{table}[ht]
    \centering
    \renewcommand{\arraystretch}{1.22}
    \setlength{\tabcolsep}{4pt}

    \caption{
        \textbf{Progressive training-data configurations.} The final row is OneStreamer. OmniMMI averages five tasks without asr input. OVOBench and StreamingBench use caption-memory inference throughout. Avg.\ is the unweighted mean of the five benchmark scores.
    }
    
    \resizebox{\textwidth}{!}{%
    \begin{tabular}{@{}ccccc|ccccc|c@{}}
        \toprule
        \textbf{Base} & \textbf{TRA-QA} & \textbf{FPPR} & \textbf{CQTC} & \textbf{SPM} &
        \textbf{OVOBench} &
        \textbf{StreamingBench} &
        \textbf{ProactiveVQA} &
        \textbf{OmniMMI} &
        \textbf{OVO-Timing} &
        \textbf{Avg.} \\
        \midrule
        $\checkmark$ & & & & & 68.3 & 85.2 & 47.5 & 22.4 & 25.7 & 49.8 \\
        $\checkmark$ & $\checkmark$ & & & & 69.4 & 85.7 & \textbf{49.7} & 26.6 & 29.4 & 52.1 \\
        $\checkmark$ & $\checkmark$ & $\checkmark$ & & & 70.9 & 85.8 & 47.3 & 29.9 & 33.0 & 53.4 \\
        $\checkmark$ & $\checkmark$ & $\checkmark$ & $\checkmark$ & & 71.1 & 85.3 & 49.0 & \textbf{30.8} & \textbf{42.0} & 55.6 \\
        \rowcolor[HTML]{DCEEFC}
        $\checkmark$ & $\checkmark$ & $\checkmark$ & $\checkmark$ & $\checkmark$ & \textbf{72.1} & \textbf{86.9} & 48.7 & 30.6 & 41.6 & \textbf{56.0} \\
        \bottomrule
    \end{tabular}%
    }
    \label{tab:data_cumulative}
\end{table}

We compare five cumulative training-data configurations that progressively introduce the key OneStreamer-Core components examined in this ablation. All five models are trained independently from the same initialization under an identical training recipe, with only the training-data composition varying across configurations. Base contains the open-source training data collected for OneStreamer-1M together with early foundational data from OneStreamer-Core. Temporal Reasoning and Anomaly QA (TRA-QA) adds data for action prediction, state grounding, multi-turn dependencies, and anomaly question answering. Fine-grained Perception and Proactive Response (FPPR) further adds continuous description, active counting, procedural QA, counting QA, and person identification. Caption Quality and Timing Calibration (CQTC) refines existing streaming caption and QA data rather than adding new samples by applying content verification and response-time calibration. Streaming Perception and Memory (SPM) finally adds general video QA, spatial QA, answerability judgment, procedural QA, and additional counting QA, producing the complete OneStreamer training mixture.

As shown in the table~\ref{tab:data_cumulative}, adding TRA-QA produces broad gains across both understanding and proactive-response benchmarks. It improves OVOBench from 68.3 to 69.4, ProactiveVQA from 47.5 to 49.7, OmniMMI from 22.4 to 26.6, and OVO-Timing from 25.7 to 29.4, raising the five-benchmark average from 49.8 to 52.1. FPPR further improves OVOBench to 70.9 and OmniMMI to 29.9, increasing the average to 53.4. Without increasing the number of training samples, CQTC yields the largest stage-wise gain on OVO-Timing, from 33.0 to 42.0, while also raising ProactiveVQA from 47.3 to 49.0 and OmniMMI from 29.9 to 30.8. The average consequently increases from 53.4 to 55.6, showing that finer content verification and response-time calibration can substantially improve the effectiveness of existing training data. Finally, SPM raises OVOBench from 71.1 to 72.1 and StreamingBench Real-Time from 85.3 to 86.9, producing the highest five-benchmark average of 56.0. The best individual proactive-response scores occur at different intermediate stages, while the full training mixture achieves the strongest aggregate performance. These results indicate that the different OneStreamer-Core components provide complementary benefits across perception, memory, and proactive response rather than uniform gains on every benchmark.

\section{Qualitative Cases}

\subsection{Memory and Perception}

\begin{figure}[ht]
    \centering
    \includegraphics[width=\textwidth,height=0.82\textheight,keepaspectratio]{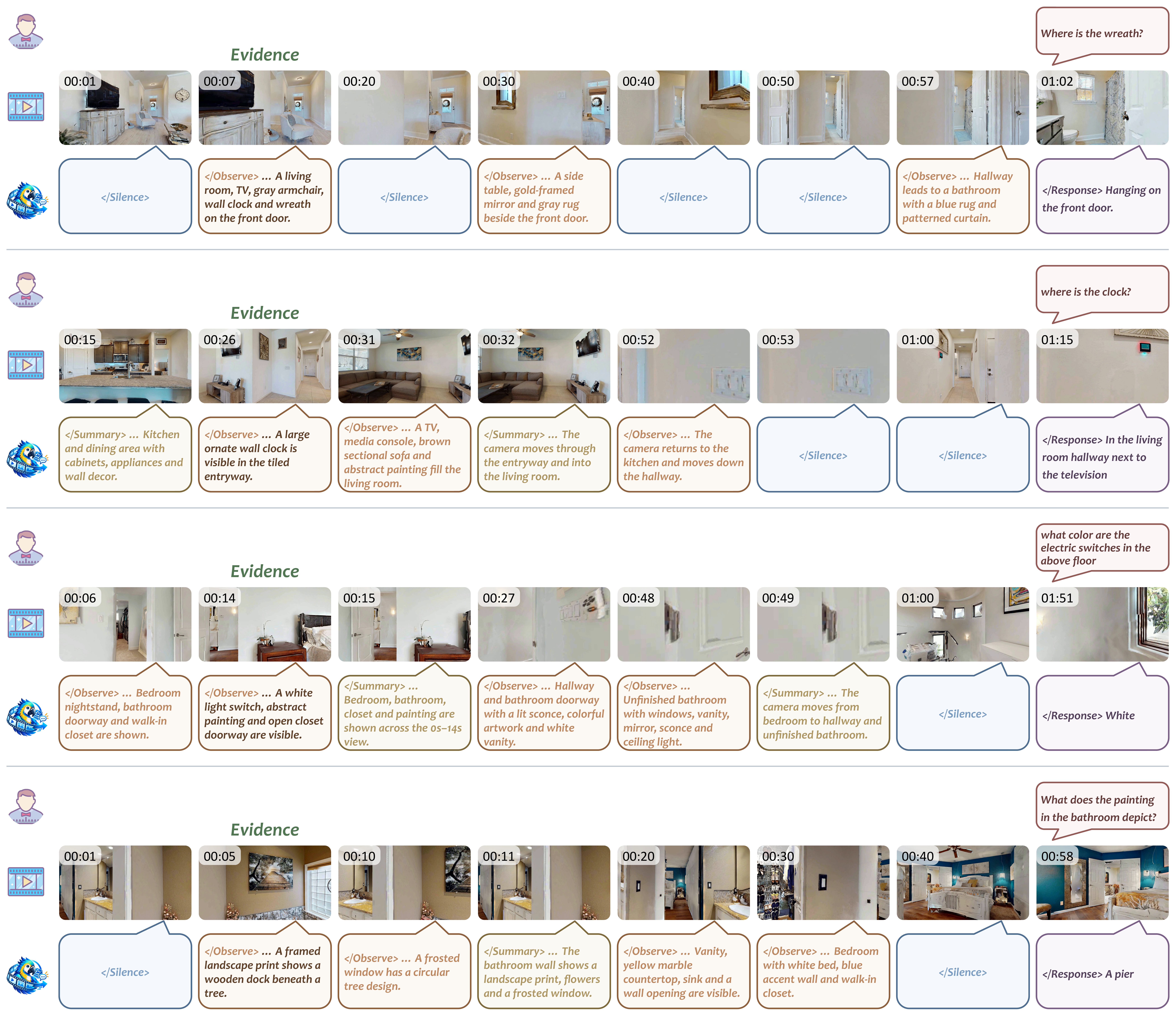}
    \caption{\textbf{Qualitative examples of memory and perception.} The model accumulates time-aligned \texttt{</Observe>} and \texttt{</Summary>} records from the stream and uses the retained evidence to answer questions about earlier observations.}
    \label{fig:app:memory_perception}
\end{figure}

Figure~\ref{fig:app:memory_perception} illustrates how {\ours} turns observations made before a question arrives into reusable factual context. As the camera moves through indoor scenes, \texttt{</Observe>} records preserve object locations, visual attributes, and artwork content. Interleaved \texttt{</Summary>} records retain broader room context and scene transitions. The resulting caption memory provides earlier evidence alongside the recent visual window when a later question is asked.

In the first example, the wreath on the front door is recorded at 00:07 and recalled when the user asks about its location at 01:02. The third example similarly records a white light switch at 00:14 and answers the later color question at 01:51. The clock example retains the earlier clock observation together with surrounding entryway and living-room context. The bathroom-painting example reuses a description of a wooden dock to answer ``A pier.'' Together, these cases illustrate how fine-grained visual observations become reusable factual memory through query-independent recording followed by task-conditioned use.

\subsection{Proactive Response}

\begin{figure}[ht]
    \centering
    \includegraphics[width=\textwidth,height=0.82\textheight,keepaspectratio]{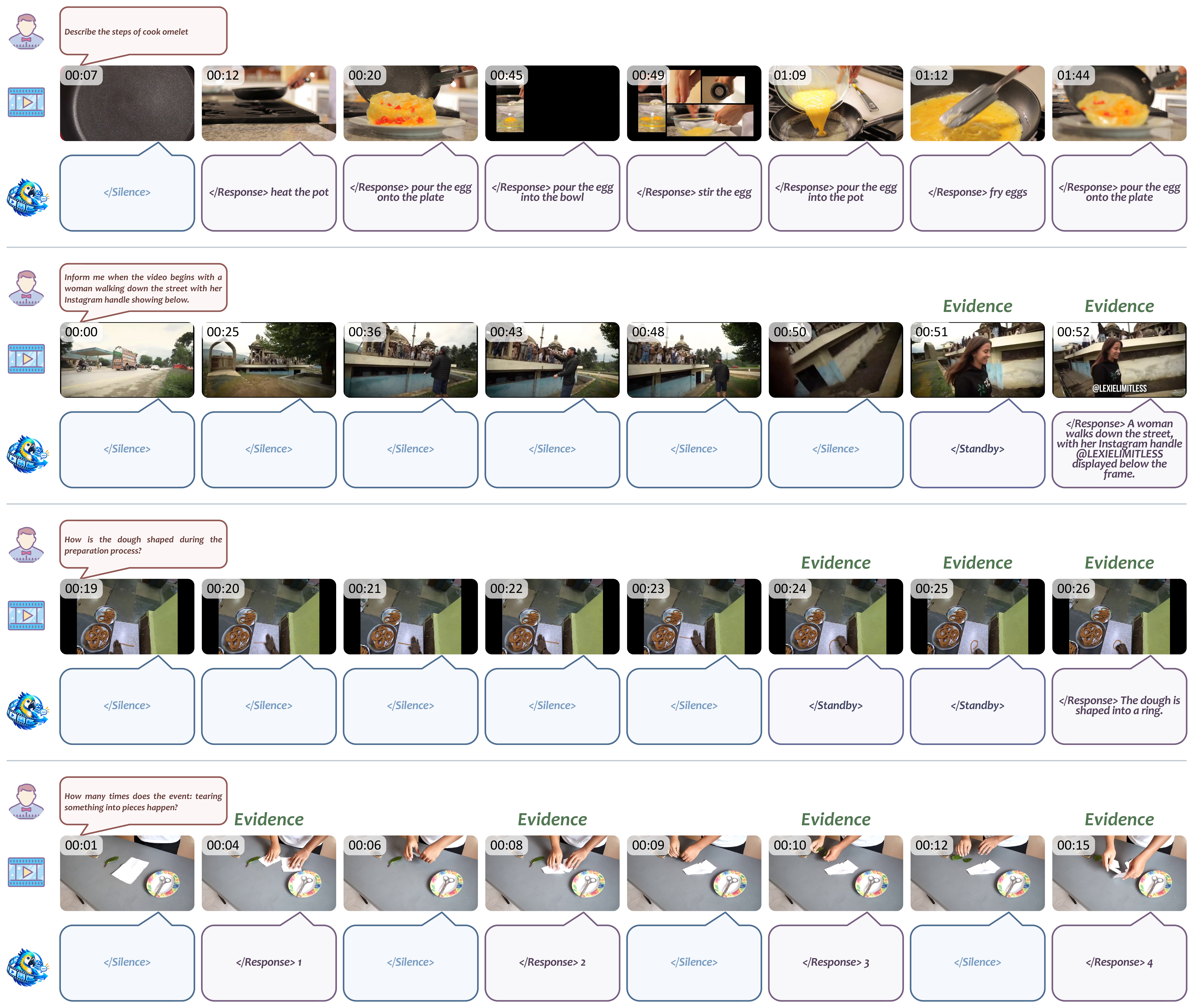}
    \caption{\textbf{Qualitative examples of proactive response.} The model remains silent while the available evidence is insufficient, enters \texttt{</Standby>} as the target event develops, and emits \texttt{</Response>} when the evidence supports the requested output.}
    \label{fig:app:proactive_response}
\end{figure}

Figure~\ref{fig:app:proactive_response} illustrates task-dependent response timing across continuous description, event notification, proactive QA, and active counting. The two middle examples show how the model waits while the requested evidence develops. In the notification example, it enters \texttt{</Standby>} at 00:51 and emits \texttt{</Response>} at 00:52, when the walking scene and the requested on-screen handle appear together. In the dough example, \texttt{</Standby>} persists from 00:24 to 00:25 before the model answers at 00:26 that the dough is shaped into a ring. These trajectories distinguish emerging evidence from evidence sufficient to support the requested output.

The top and bottom examples require repeated outputs under a single instruction. In the cooking example, {\ours} describes observed actions incrementally as the video progresses. In the counting example, it updates the cumulative number of tearing events from one to four at 00:04, 00:08, 00:10, and 00:15, with \texttt{</Silence>} between updates. Together, these cases show how response-control states support both continued observation and successive task outputs as the relevant evidence becomes available.

\subsection{Failure Cases}

\begin{figure}[ht]
    \centering
    \includegraphics[width=\textwidth]{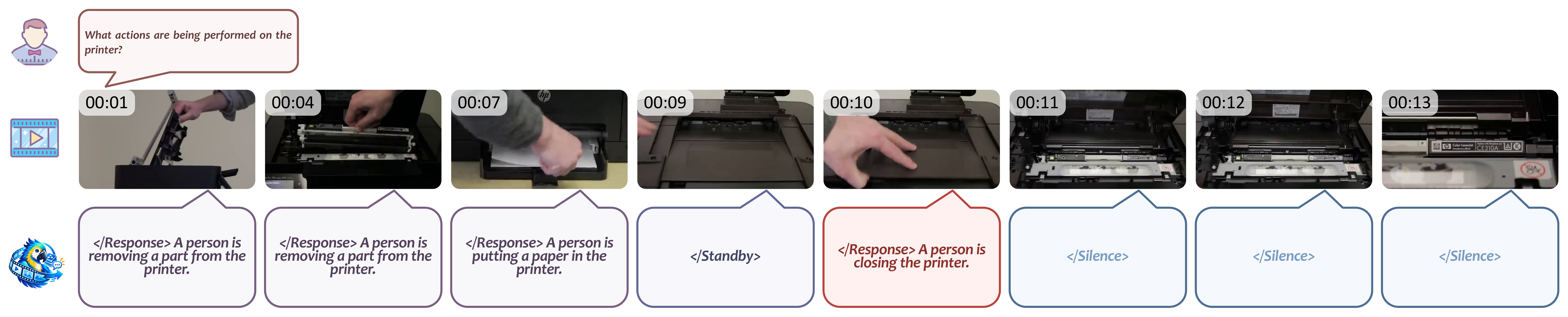}
    \caption{\textbf{Failure case in proactive action description.} At 00:10, the model incorrectly reports that the printer is being closed, as highlighted in red.}
    \label{fig:app:failure_cases}
\end{figure}

Figure~\ref{fig:app:failure_cases} illustrates a failure in action interpretation followed by a lack of correction. Asked to describe the actions performed on a printer, {\ours} produces incremental responses as the video unfolds. It enters \texttt{</Standby>} at 00:09 and emits \texttt{</Response>} at 00:10 with the description ``A person is closing the printer.'' However, the displayed sequence shows the cover being lifted to expose the printer interior rather than closed. The subsequent frames at 00:11--00:13 make the open state clear, yet the model outputs \texttt{</Silence>} at all three timestamps and does not revise its earlier description within the shown sequence. This case highlights two remaining challenges for evidence-grounded streaming interaction: accurately interpreting object-state changes and correcting earlier responses when subsequent evidence reveals an error.

\end{document}